\documentclass[10pt,twocolumn,letterpaper]{article}

\usepackage{cvpr}      
\definecolor{cvprblue}{rgb}{0.21,0.49,0.74}
\usepackage[pagebackref,breaklinks,colorlinks,allcolors=cvprblue]{hyperref}
\usepackage{multirow}
\def\paperID{2835} 
\def\confName{CVPR}
\def\confYear{2026}

\title{MERLIN: Building Low-SNR Robust Multimodal LLMs for Electromagnetic Signals\thanks{Project page: \url{https://em-merlin.github.io}}}

\author{
  Junyu Shen$^{1\ast}$,
  Zhendong She$^{3\ast}$,
  Chenghanyu Zhang$^{2}$\thanks{Contribute equally to this work.},
  Yuchuang Sun$^{1}$,
  Luqing Luo$^{4}$, \\
  Dingwei Tan$^{5}$,
  Zonghao Guo$^{1}$,
  Bo Guo$^{1}$,
  Zehua Han$^{7}$,
  Wupeng Xie$^{9}$,
  Yaxin Mu$^{8}$, \\
  Peng Zhang$^{3}$,
  Peipei Li$^{2}$,
  Fengxiang Wang$^{6\dagger}$,
  Yangang Sun$^{1\dagger}$,
  Maosong Sun$^{1}$\thanks{Corresponding author.}
  \vspace{2mm} 
  \\
  $^1$ Tsinghua University \quad
  $^2$ Beijing University of Posts and Telecommunications \\
  $^3$ Tianjin University \quad
  $^4$ Institute of Microelectronics of the Chinese Academy of Sciences \\
  $^5$ HKUST (Guangzhou) \quad
  $^6$ National University of Defense Technology \\
  $^7$ Beihang University \quad
  $^8$ Beijing Information Science and Technology University \\
  $^9$ Artificial Intelligence Institute of China Electronics Technology Group Corporation \\
}

\usepackage{booktabs}
\usepackage{multirow}
\usepackage{graphicx}
\usepackage[table]{xcolor} 
\usepackage{caption}
\let\oldtwocolumn\twocolumn
\renewcommand\twocolumn[1][]{%
    \oldtwocolumn[{#1}{
    \begin{center}
    \captionsetup{type=figure}
    \includegraphics[width=\textwidth]{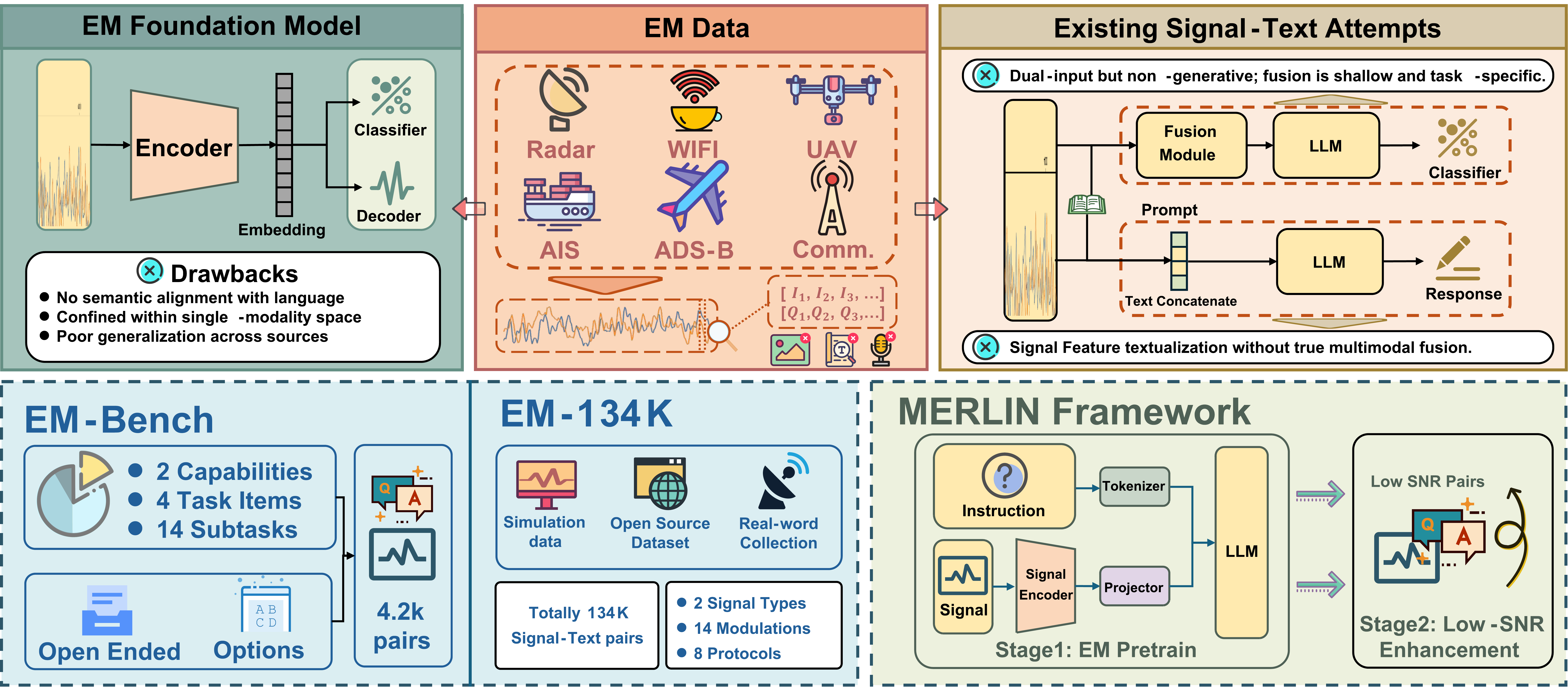}
    \captionof{figure}{Overview of our work. The top panel highlights the challenge of applying general MLLMs to the unique domain of EM signals. The bottom panel outlines our contributions to solve this: the EM-Bench benchmark and EM-100K dataset to address data scarcity, and the MERLIN framework, which employs multitask pre-training and low-SNR self-adaptation for robust performance.}
    \label{fig:teaser}
  \end{center}
    }]
}

\begin{document}
\maketitle


\begin{abstract}
    The paradigm of Multimodal Large Language Models (MLLMs) offers a promising blueprint for advancing the electromagnetic (EM) domain. However, prevailing approaches often deviate from the native MLLM paradigm, instead using task-specific or pipelined architectures that lead to fundamental limitations in model performance and generalization. Fully realizing the MLLM potential in EM domain requires overcoming three main challenges: (1) \textbf{Data.} The scarcity of high-quality datasets with paired EM signals and descriptive text annotations used for MLLM pre-training; (2) \textbf{Benchmark.} The absence of comprehensive benchmarks to systematically evaluate and compare the performance of models on EM signal-to-text tasks; (3) \textbf{Model.}
    A critical fragility in low Signal-to-Noise Ratio (SNR) environments, where critical signal features can be obscured, leading to significant performance degradation. 
    To address these challenges, we introduce a tripartite contribution to establish a foundation for MLLMs in the EM domain. First, to overcome data scarcity, we construct and release EM-134k, a large-scale dataset comprising over 134,000 EM signal-text pairs. Second, to enable rigorous and standardized evaluation, we propose EM-Bench, the most comprehensive benchmark featuring diverse downstream tasks spanning from perception to reasoning. Finally, to tackle the core modeling challenge, we present MERLIN, a novel training framework designed not only to align low-level signal representations with high-level semantic text, but also to explicitly enhance model robustness and performance in challenging low-SNR environments. Comprehensive experiments validate our method, showing that MERLIN is state-of-the-art in the EM-Bench and exhibits remarkable robustness in low-SNR settings.

\end{abstract}

\section{Introduction}
\label{sec:intro}

The precise perception and reasoning of features within electromagnetic (EM) signals is crucial for critical domains such as radar, communications, and navigation~\cite{davaslioglu2022self, kim2024vit, kang2024sensing,quan2025signal}. Although deep learning has shown success in basic EM perception tasks, most existing methods are designed for task-specific applications. They generally lack the ability to generalize across a diverse range of perception and reasoning challenges, limiting their potential in real-world scenarios.

Recent advancements in Multimodal Large Language Models (MLLMs)~\cite{ahmed2025qwen, achiam2023gpt, team2023gemini, zhu2023minigpt, bai2023qwen,liu2023visual,li2024llava} offer a promising path toward multi-task generalization in the EM domain~\cite{chen2025radiollm, cheng2025large}. 
By bridging raw signals with the versatile reasoning capabilities of LLMs, these models could unlock unprecedented potential. Early efforts to achieve joint understanding of electromagnetic signals and natural language have laid important groundwork. However, these pioneering works often deviate from the standard, end-to-end paradigm that has proven successful for modern MLLMs in domains like vision. Many existing methods adopt a pipelined or task-specific architecture. Consequently, their generalize performance is often capped. Therefore, adapting a native MLLM framework to the EM domain presents a promising and underexplored direction. However, a direct application of existing vision-language paradigms is impeded by three fundamental challenges unique to the electromagnetic domain:



\textbf{Data Scarcity.} The inherent confidentiality and complexity of EM signals have led to an extreme lack of high-quality, large-scale public datasets pairing signals with text. This data bottleneck fundamentally restricts the training and development of capable EM MLLMs. To address this barrier, we construct EM-134k, a large-scale dataset comprising over 134,000 signal-text pairs, integrating open-source signals, professional simulations, and real-world samples to provide a robust foundation for model pre-training.

\textbf{The Absence of Standardized Benchmarks.} Without a common ground for evaluation, it is impossible to fairly compare different model architectures and training strategies. To resolve this and enable measurable progress, we introduce EM-Bench, the first comprehensive benchmark designed to evaluate MLLMs in the EM domain. EM-Bench features a multi-tiered evaluation system covering tasks that span from fundamental perception to complex reasoning, offering over 4,200 high-quality question-answering pairs for a standardized and multi-dimensional assessment of model capabilities.


\textbf{Model Performance Degradation in Low-SNR Environments.} Enhancing robustness in noisy conditions remains a central challenge in the EM domain. We identify a critical fragility in the standard encoder-LLM architecture: its performance collapses dramatically in low Signal-to-Noise Ratio (SNR) environments. In this work, we define low-SNR as any signal with an SNR below 0 dB, a critical threshold where the noise power exceeds the signal power. This failure occurs because noise corrupts the low-level signal features~\cite{jiang2023radiation,chen2023noise, weng2025self,oh2025multi}, exacerbating the semantic gap between the signal and text modalities. Overcoming this fragility is not merely an incremental improvement, but a core challenge that must be solved to make EM MLLMs viable for practical application. Notably, our experiments demonstrate that the model tends to extract homogenized features, thereby losing the ability to discriminate the intrinsic physical representation. We furthermore observe a key phenomenon: linearly interpolating from the feature of a noisy signal towards its clean version systematically improves model performance. These findings, combined with insights from advanced low-SNR handling techniques in fields such as image enhancement~\cite{xu2022snr,makwana2024livenet,zhang2024adaptive} and EEG signal representation learning~\cite{wang2024eegpt, basheer2024improving, basheer2024improving, im2025cocl}, inspired our conclusion: tackling noise directly at the feature level is an effective pathway.

To this end, we propose \textbf{MERLIN} (Multi-modal Electromagnetic Robust Learning), a two-stage training framework designed to build an EM MLLM robust to low-SNR conditions. Following an initial multi-task pre-training stage, MERLIN introduces a second stage based on knowledge distillation\cite{lyu2024knowtuning, liu2024evolving, peng2025enhancing}. In this stage, the pre-trained model initializes a frozen teacher network and a trainable student network. The student is trained on low-SNR signals to align its latent representations with those of the teacher processing the corresponding high-SNR signals. This process compels the student to learn a noise-invariant representation, significantly improving performance in noisy conditions while preserving its general multi-task capabilities.

In summary, our core contributions are as follows:
\begin{itemize}
    \item We built EM-134K, a large-scale pre-training dataset for EM signals, and introduced EM-Bench, the field's most comprehensive evaluation benchmark. These resources address the critical bottlenecks of data scarcity and the lack of standardized evaluation for multimodal EM models, providing essential infrastructure for future research.
    \item We propose MERLIN, a novel two-stage training framework designed to build EM MLLM robust to low-SNR conditions. By integrating an initial pre-training stage with a subsequent knowledge distillation stage for low-SNR robustness, MERLIN produces a model that excels in noisy environments while retaining its foundational multi-task abilities. This method effectively resolves the key challenge of low-SNR performance collapse.
    \item We conduct extensive evaluations on our proposed EM-Bench. Our MERLIN-trained model demonstrates remarkable robustness in low-SNR settings and substantially outperforms baselines. This result validates the efficacy of our end-to-end multimodal method.
\end{itemize}

\section{Related Work}
\label{sec:rw}
\textbf{Multimodal Large Language Models.} The powerful reasoning of LLMs, exemplified by PaLM-2~\cite{anil2023palm}, is now being extended to perceive and understand non-textual information through MLLMs. This wave of innovation began in the visual domain, where a series of influential models learned to align image features with language to enable complex visual dialogue~\cite{li2023blip, liu2023visual, wang2024qwen2, li2024llava, achiam2023gpt, comanici2025gemini, chen2024internvl}. This core idea was quickly adapted to interpret temporal dynamics in videos~\cite{alayrac2022flamingo, zhang2023video, li2023videochat, yan2021videogpt, lin2024video} and to process raw audio for end-to-end speech interaction~\cite{rubenstein2023audiopalm, zhang2023speechgpt, chen2024vall, tang2023salmonn}. More recently, the electromagnetic domain has emerged as a new frontier for this multimodal approach. Pioneering works in this area aim to bridge low-level signal data with high-level semantic understanding. For instance, RadioLLM~\cite{chen2025radiollm} leverages a Q-Former architecture, popularized by vision models like BLIP-2~\cite{li2023blip}, to map I/Q signal features into the LLM's semantic space. Similarly, Spectrum-LLM~\cite{liu2025spectrumllm} and WirelessLLM~\cite{shao2024wirelessllm} are developing novel encoding strategies to translate time-series signals and network data into formats digestible by LLMs, paving the way for natural language-based wireless cognition.

\textbf{Electromagnetic Datasets.} The development of robust wireless perception models heavily relies on high-quality datasets. The RadioML series~\cite{o2016radio,o2018over} established a foundational benchmark for modulation classification under varying SNR conditions. Subsequent datasets, including HisarMod~\cite{tekbiyik2019hisarmod} and RadarCommDataset~\cite{jagannath2021dataset}, expanded the scope by providing a greater diversity of modulation schemes, channel conditions, and signal protocols. More recently, the creation of large-scale corpora like EM\cite{luo2025emind}, with its 800,000 raw I/Q samples, has provided the necessary infrastructure for pre-training large, generalizable models for the wireless domain.

\textbf{Low-SNR Robust Perception.} The pervasive low signal-to-noise ratio (SNR) problem fundamentally constrains the feasibility and performance ceilings of perception and reasoning tasks, making robustness enhancement a critical challenge across multiple domains. In the image domain, researchers have developed various noise-robust methodologies, including physics-guided enhancement frameworks \cite{xu2022snr, taassori2025robustdeit, tao2025nevlp, shi2024survey}, adaptive multi-scale attention mechanisms \cite{makwana2024livenet,zhou2024adapt, ahn2025integrating, khudjaev2024dformer}, and contrastive learning-driven feature recovery techniques \cite{oh2025multi, zhang2024adaptive,li2024noisy,jebur2024comprehensive, jovhari2025noise}, significantly improving visual perception capabilities under low-light and high-noise conditions. 

\section{EM-134K \& EM-Bench}
\label{sec:dataset}

\begin{table*}[ht]
    \caption{Composition of the large-scale data corpus used for creating EM-134K and EM-Bench. This table details the 10 subsets, totaling over 35 million signal samples, that form our foundational data pool.}
    \label{tab:dataset}
    \centering
    \resizebox{\textwidth}{!}{
    \begin{tabular}{lccccc}
        \toprule
        Subset & Signal Type & Mission Type & Data Source & Number of Samples\\
        \midrule
        MOD & -- & 14 types of modulation & Real-world & 588,000 \\
        PE & LFM/Barker/Frank/Stepped/Rectangular & 6 types of parameters & Simulation & 7,310,600 \\  
        PI & BPSK/QPSK/16-QAM/64-QAM/256-QAM & 8 types of protocols & Simulation & 911,895 \\
        SD & -- & 9 types of Jamming & Simulation & 162,000 \\
        RJR & LFM/Barker/Noise/Frank/Sin/Rectangular & 12 types of Jamming & Simulation & 4,918,700 \\
        CJR & BPSK/QPSK/16-QAM/64-QAM/GFSK & 9 types of Jamming & Simulation & 4,984,875 \\
        RJS & LFM/Barker/Frank/Stepped/Rectangular & -- & Simulation & 5,313,600 \\
        CJS & -- & -- & Simulation & 2,747,000 \\
        Anti-RJ & LFM/Barker/Frank/Noise/Rectangular & -- & Simulation & 3,948,200 \\
        Anti-CJ & BPSK/QPSK/16-QAM/64-QAM/GFSK & -- & Simulation & 4,984,875 \\
        \midrule
        TOTAL & & & & 35,869,745 \\
        \bottomrule
    \end{tabular}
    }
\end{table*}

\subsection{Evaluation Dimensions}
Performing an effective analysis of EM signals requires two fundamental abilities. The first is perception, the ability to understand the core characteristics of raw signal data. The second is reasoning, used to draw advanced conclusions and develop strategies. For MLLMs to perform well in complex EM scenarios, they must effectively combine robust perception and reasoning. Therefore, we structure the EM-Bench evaluation around these two primary (\textbf{L-1}) capabilities: Perception and Reasoning. These are further divided into 4 \textbf{L-2} tasks and then refined into 14 fine-grained \textbf{L-3} sub-tasks, shown in Figure \ref{fig:evaluation_framework}. Perception covers 3 L-2 tasks and is mainly evaluated using single-choice questions. In contrast, reasoning covers 1 L-2 task and is assessed via open-ended question answering. The details of each L-2 task are presented as follows:

\begin{figure}[ht]
\centering
\includegraphics[width=\columnwidth]{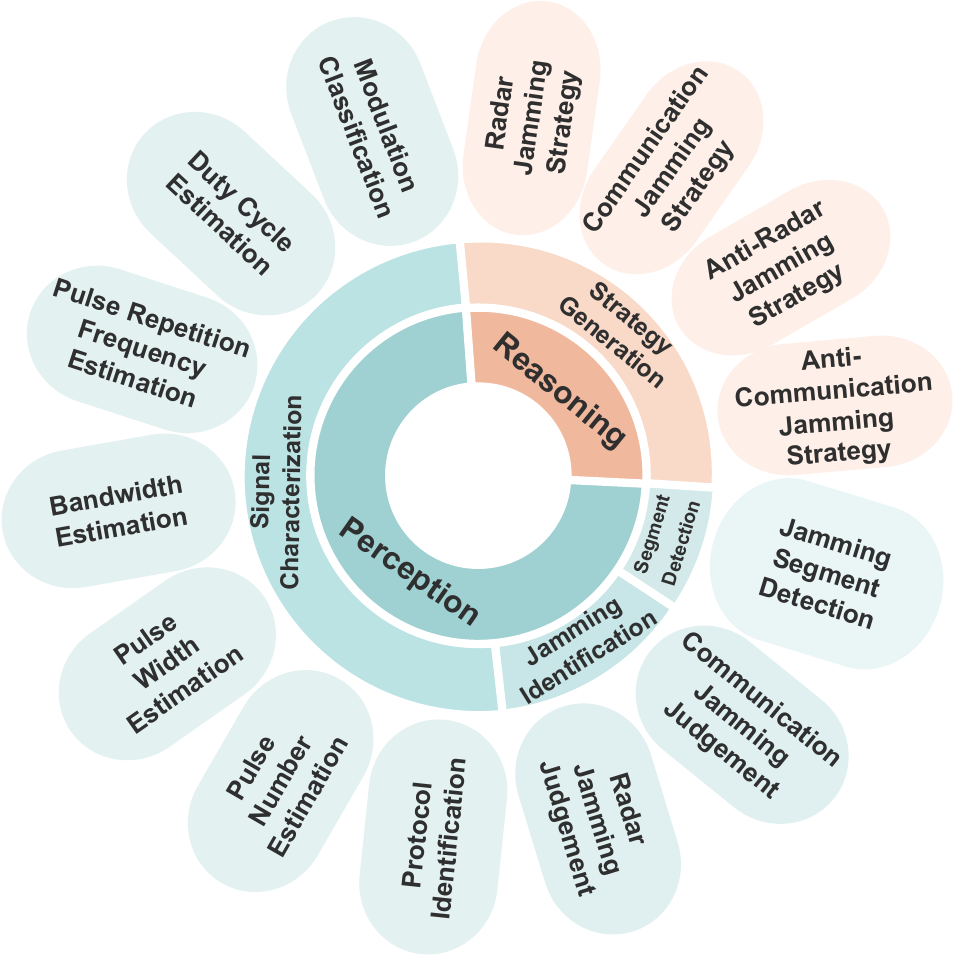}
\caption{The hierarchical evaluation framework of EM-Bench, which systematically assesses the perception and reasoning capabilities of MLLMs on electromagnetic IQ signals across \textit{3} levels and \textit{14} sub-tasks.}
\label{fig:evaluation_framework}
\end{figure}

\begin{itemize}[left=0pt]
    \item \textbf{Signal Characterization.} Electromagnetic IQ signals encode rich information, including modulation, protocol, and key parameter estimations. This task is therefore composed of 8 L-3 sub-tasks, each designed to evaluate the model's ability to identify these distinct attributes. To meticulously assess the MLLM's precision in parameter perception, we extract all 6 key parameters from the parameter estimation tasks in EM-134K, formulating a dedicated sub-task for each.
    \item \textbf{Jamming Identification.} Signal jamming is a critical challenge in the EM domain, as strong interference can severely degrade signal quality. To evaluate an MLLM's ability to perceive interference, we design 2 L-3 sub-tasks covering 12 typical jamming types in the radar domain and 9 in the communications domain.
    \item \textbf{Fragment Detection.} Beyond identifying the type of jamming, an effective MLLM also needs the ability to localize the affected signal segments. This sub-task requires MLLMs to output the start and end points of the interfered IQ signal segment, thereby evaluating its fine-grained temporal perception capabilities.
    \item \textbf{Strategy Generation.} In scenarios such as electronic countermeasures, MLLMs can serve as assistants in devising strategies to jam hostile signals or to counteract adversarial jamming. This L-2 task aims to evaluate the MLLM's capacity for strategic planning. It comprises 4 L-3 sub-tasks, corresponding to the generation of jamming and anti-jamming strategies in both radar and communication contexts.
\end{itemize}

\subsection{Data Collection and QA Pairs Construction}
\begin{figure}[ht]
\centering
\includegraphics[width=\columnwidth]{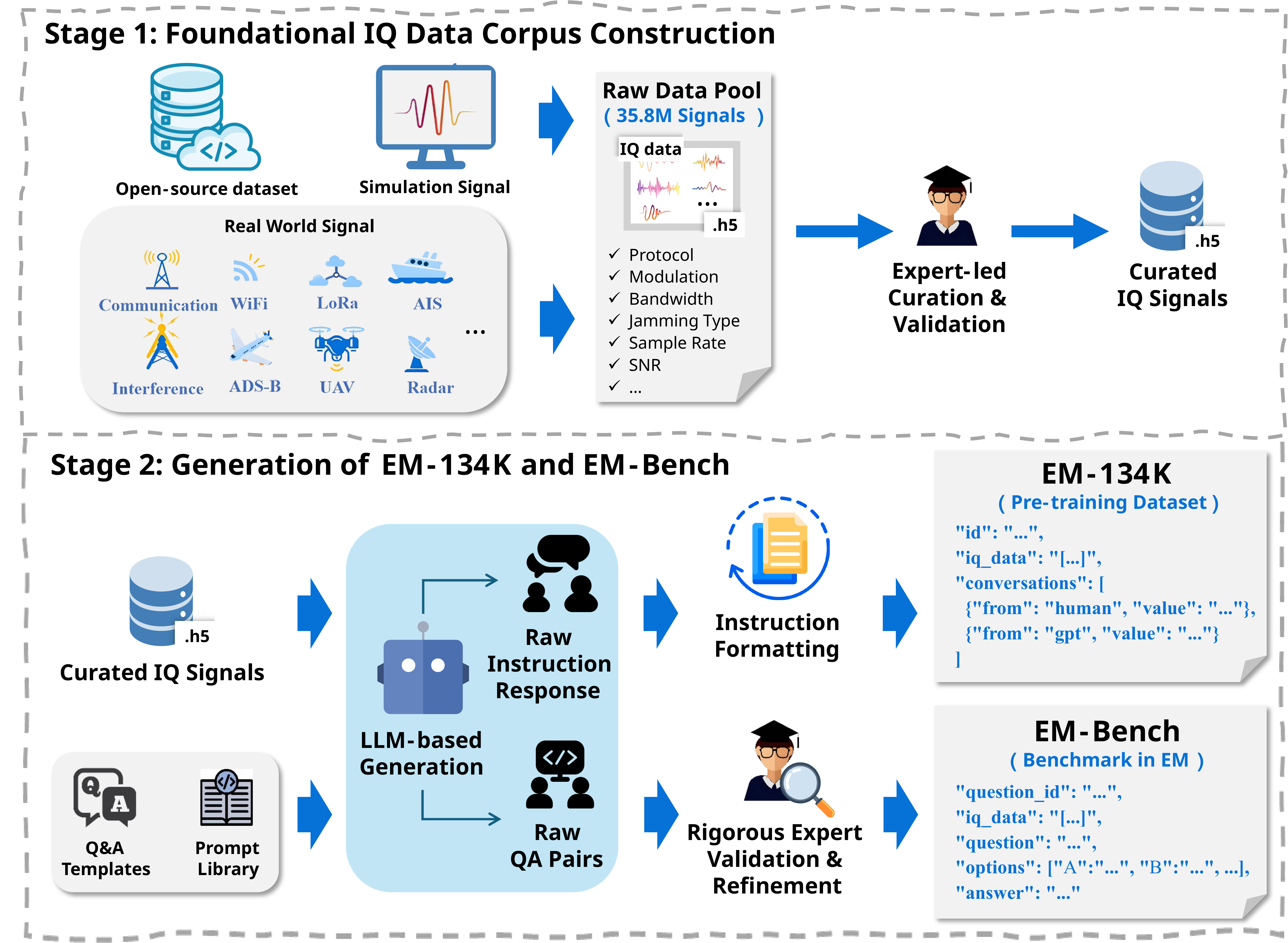} 
\caption{The two-stage construction pipeline for the EM-134K fine-tuning dataset and the EM-Bench evaluation benchmark. Stage 1 focuses on building a large-scale, expert-validated IQ data corpus. Stage 2 details the distinct generation processes: a direct, large-scale formatting for EM-134K, and a rigorously expert-validated pathway for the high-quality EM-Bench.}
\label{fig:data-pipeline}
\end{figure}

\textbf{Data Sourcing.} To construct a diverse and transferable multi-task dataset, we focus on maximizing the diversity of both the EM signals and their associated tasks. As illustrated in Figure ~\ref{fig:data-pipeline} (Stage 1). We collect a diverse corpus of IQ data by synthetic generation and real-world collection. Furthermore, we integrate an existing open-source dataset~\cite{huang2023multi} to ensure comprehension in the parameter estimation task. As summarized in Table \ref{tab:dataset}, our final dataset contains 35,869,745 IQ signals across 10 distinct sub-tasks, covering a wide range of signal types and parameters. All signals are standardized to a 20 MHz sampling rate, a uniform SNR distribution from -20 to 20 dB, and a fixed length of 1024 samples. To ensure annotation quality, we retain the original labels for public datasets, while the corresponding labels for our synthetic data are generated programmatically during the simulation process. Finally, the entire dataset is validated by domain experts who confirmed the consistency between the IQ waveforms and their parameters to confirm the fidelity.

\textbf{EM-134K \& EM-Bench Construction.} Our pipeline to create the benchmark and dataset begins with a unified, LLM-driven generation process, as shown in Figure \ref{fig:data-pipeline} (Stage 2). This core module takes the Curated IQ Signals along with our QA Templates and Prompt Library as inputs to produce a large, shared pool of raw outputs. From this common origin, the data is channeled into two distinct pathways to forge the EM-Bench and EM-134K datasets, each with a tailored procedure reflecting its end purpose.

\begin{figure*}[ht]
\centering
\includegraphics[width=\textwidth]{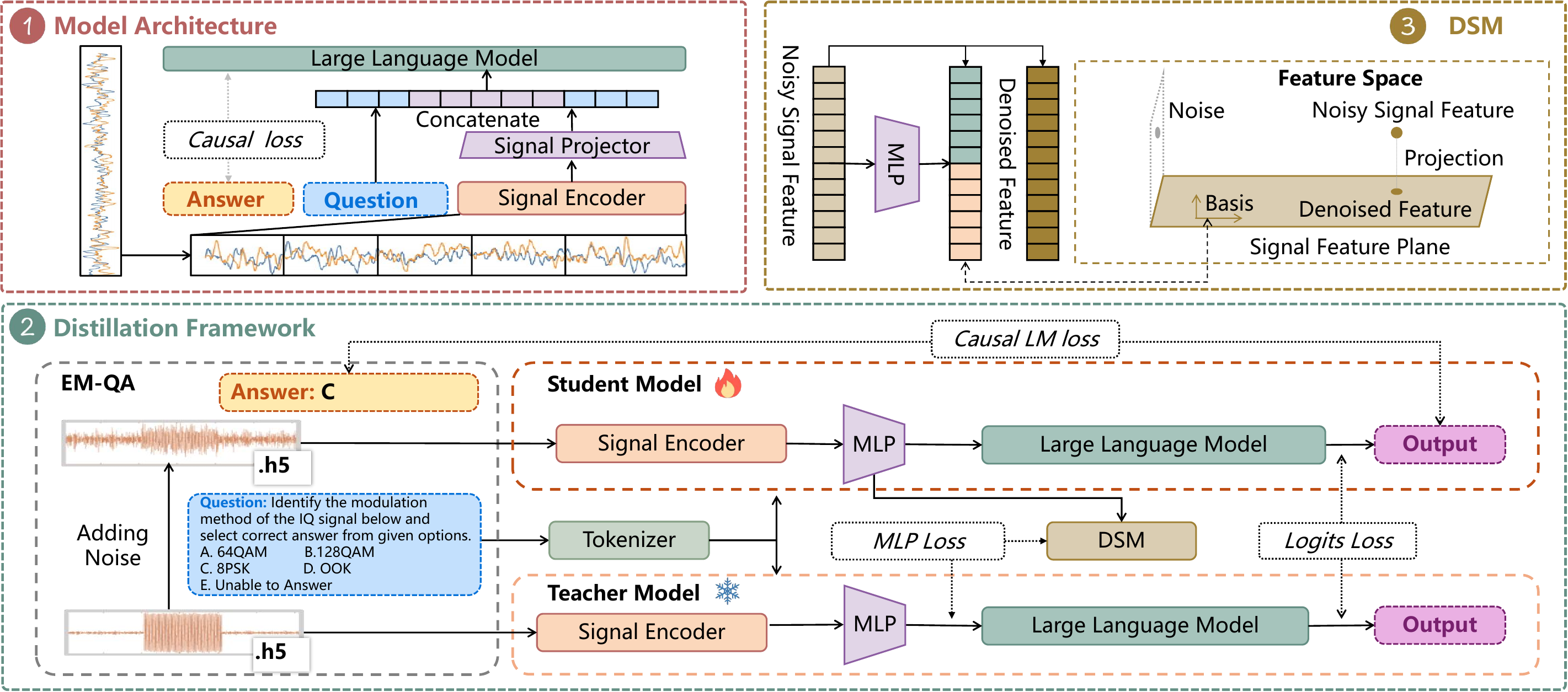} 
\caption{The architecture and training framework of MERLIN. (1) The baseline model architecture consists of a Signal Encoder, a Projector, and a LLM. (2) The knowledge distillation framework enhances low-SNR robustness by using a frozen high-SNR teacher model to guide a student model. (3) The Denoising Subspace Module (DSM) facilitates effective distillation by projecting noisy signal features into a clean, noise-invariant feature space.}
\label{fig:merlin}
\end{figure*}

To construct the EM-Bench benchmark, raw QA pairs from the generated pool are subjected to a Rigorous Expert Validation and Refinement process. This critical stage ensures the highest quality, as domain experts meticulously review each question-answer pair for factual accuracy, clarity, and logical soundness. Any ambiguous or incorrect samples are either refined or discarded.

For the EM-134K fine-tuning dataset, the process is optimized for scale. Raw instruction-response pairs are channeled through an automated Instruction Formatting pipeline. This step programmatically structures the data into a conversational format suitable for large-scale supervised fine-tuning, efficiently converting the raw LLM outputs into a massive training resource. 

\subsection{Analysis}
The dual-path construction pipeline yields two distinct resources, EM-Bench and EM-134K, whose properties and intended roles are analyzed below.

EM-Bench is a high-quality benchmark designed for comprehensive evaluation, comprising 4,200 expert-validated QA pairs. It is structured to provide 300 pairs for each of the 14 L-3 sub-tasks defined in our evaluation framework. To thoroughly assess different model capabilities, the benchmark utilizes two distinct formats:

\begin{itemize}[left=0pt]
    \item \textbf{Perception Tasks} use a single-choice format represented by the tuple $P_i=(Q_i, C_i, I_i, A_i)$, where $Q_i$ denotes the question, $I_i$ denotes the IQ signal sequence, $A_i$ denotes the ground-truth answer, and $C_i$ denotes a set of five options containing the correct answer, three distractors randomly sampled from the ground-truth answers of other QA pairs, and an option "Unable to answer" for confidence assessment.
    \item \textbf{Reasoning Tasks} follow a constrained open-ended format $P_i=(Q_i, I_i, A_i)$, which requires models to generate strategic responses directly without predefined options.
\end{itemize}

EM-134K is a large-scale dataset created for supervised fine-tuning, consisting of 134,107 instruction-tuning samples. Following a LLaVA-like architecture~\cite{liu2023visual}, each sample is formatted as a single-turn dialogue (an instruction and a ground-truth response). This structure decouples the raw IQ data from the text while retaining embedded metadata such as sampling rate and SNR, making it ideal for training.

In summary, EM-Bench provides a high-fidelity yardstick for measuring model capabilities, while EM-134K offers the large-scale, specialized knowledge needed to enhance them. Together, they form a comprehensive and synergistic ecosystem for advancing Multimodal Large Language Models in the electromagnetic domain.

\section{MERLIN Framework}
\label{sec:method}


 We introduce MERLIN, a comprehensive framework for building robust and capable EM MLLMs. To address the performance degradation in low-SNR conditions, MERLIN adopts a carefully structured two-stage training paragdigm: (1) Foundational cross-modal pretraining to establish a robust link between the two modalities, and (2) Low-SNR robustness enhancement to specifically overcome the model's fragility in noisy conditions. 

\begin{figure*}[htbp]
\centering

\begin{subfigure}[b]{0.3\textwidth}
    \centering
    \includegraphics[width=\linewidth]{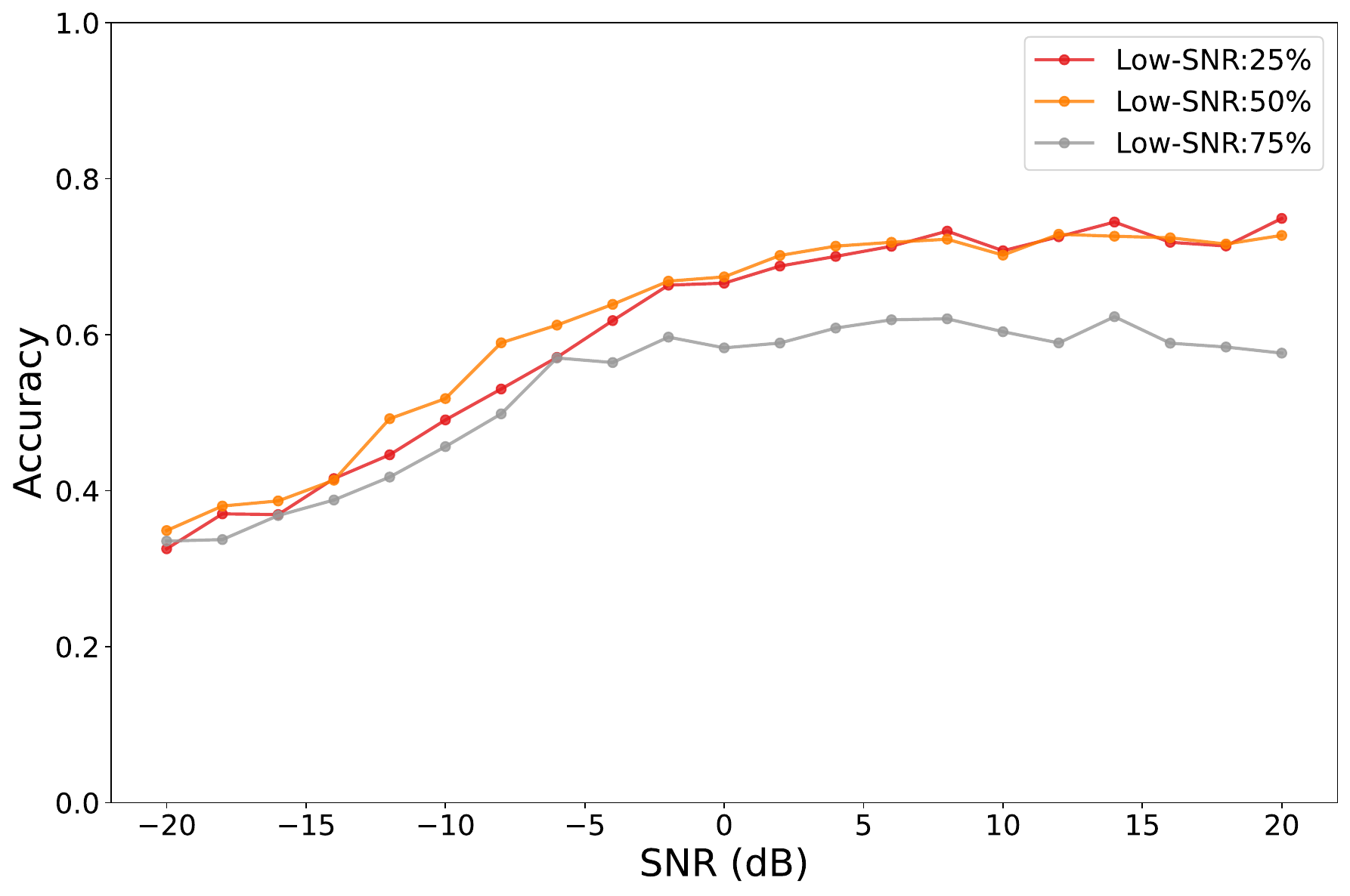}
    \caption{Comparison of training results for high and low SNR data in different mixed scenarios.}
    \label{fig:group_a}
\end{subfigure}
\hfill 
\begin{subfigure}[b]{0.3\textwidth}
    \centering
    \begin{minipage}[b]{0.48\linewidth}
        \centering
        \includegraphics[width=\linewidth]{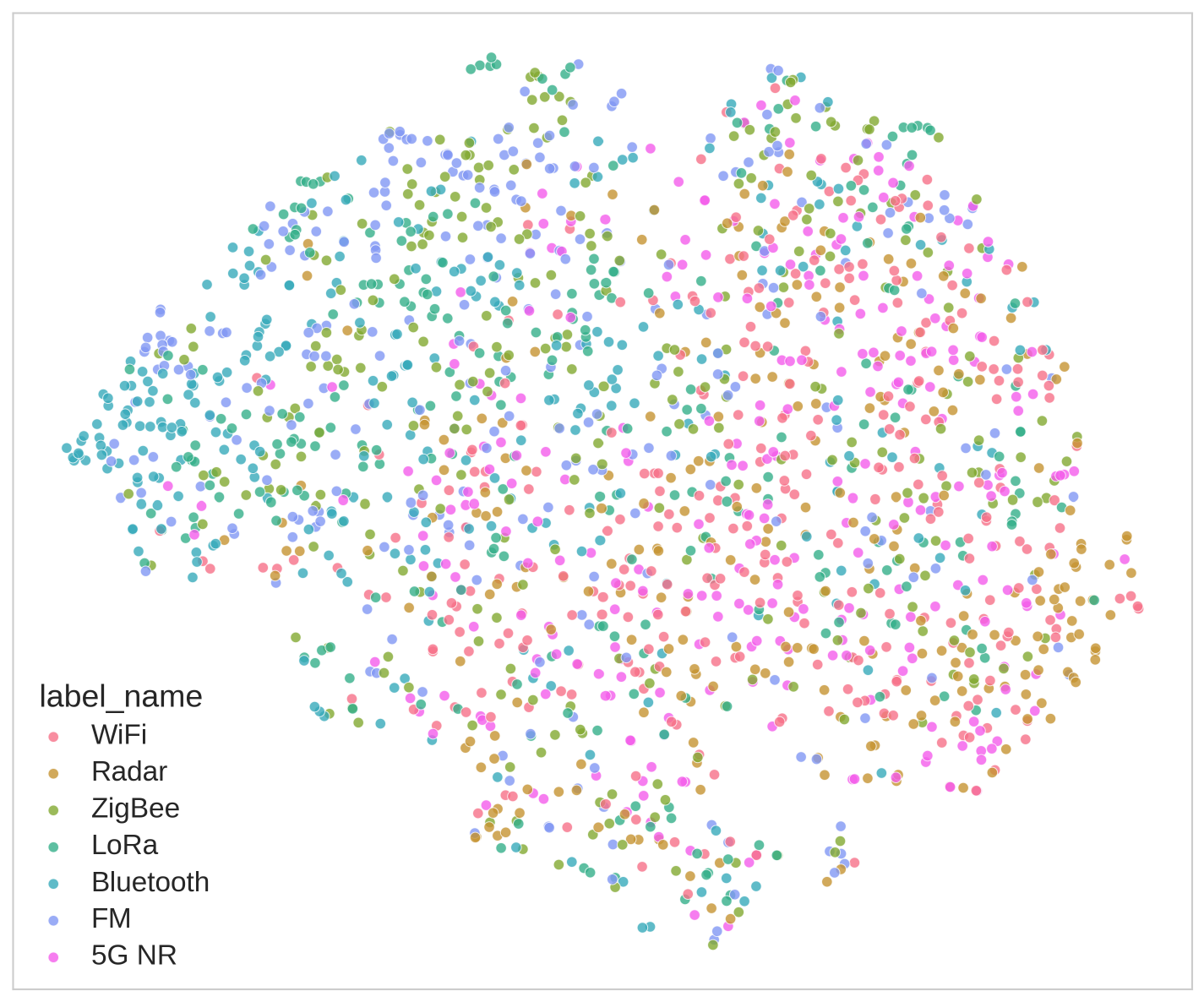}
        \small -20db to -10db
    \end{minipage}
    \hfill
    \begin{minipage}[b]{0.48\linewidth}
        \centering
        \includegraphics[width=\linewidth]{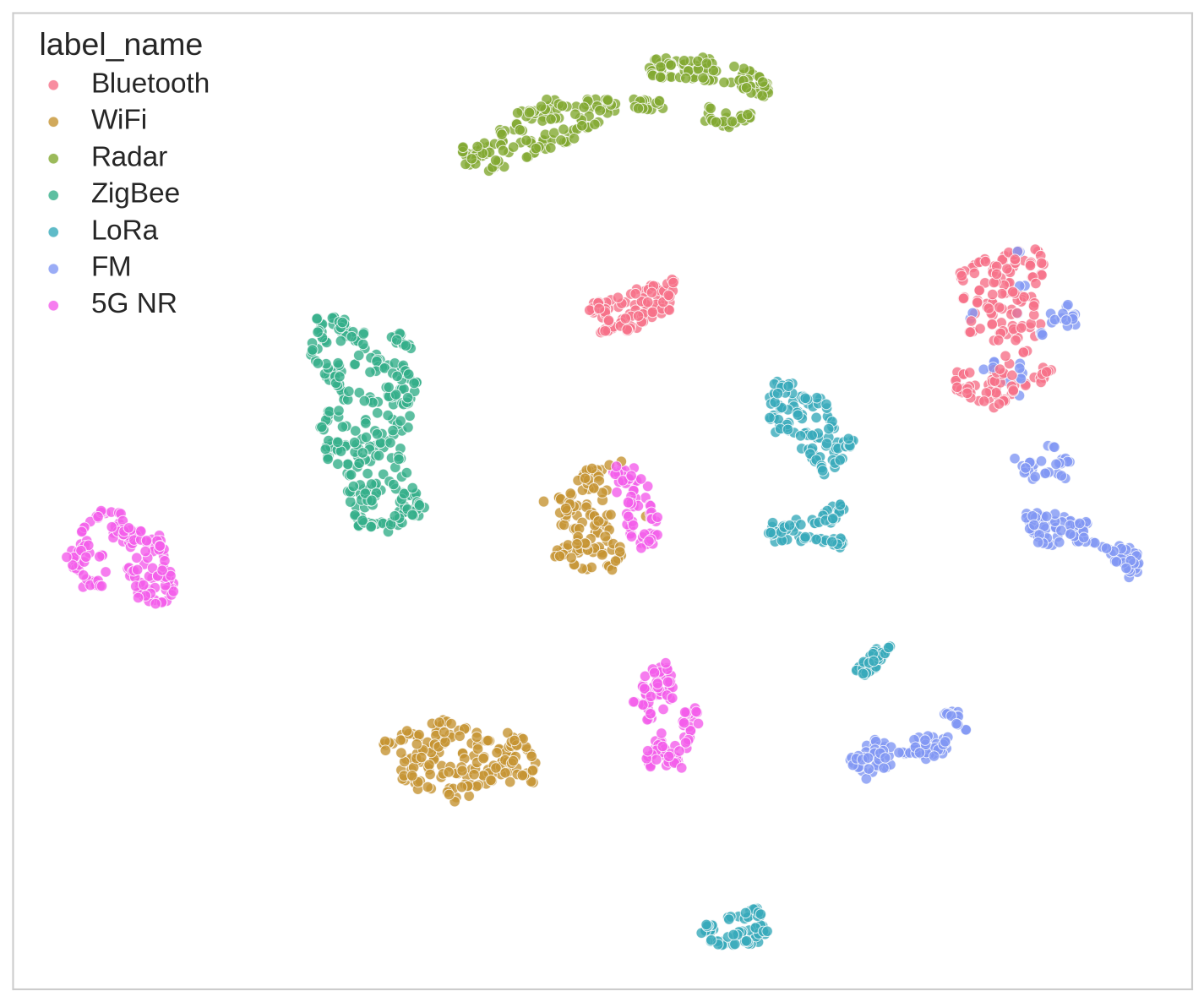}
        \small 10db to 20db
    \end{minipage}
    \vspace{2mm} 
    \caption{Embedding clustering of 2000 samples for the PI task under different SNR groups.}
    \label{fig:group_b}
\end{subfigure}
\hfill 
\begin{subfigure}[b]{0.3\textwidth}
    \centering
    \includegraphics[width=\linewidth]{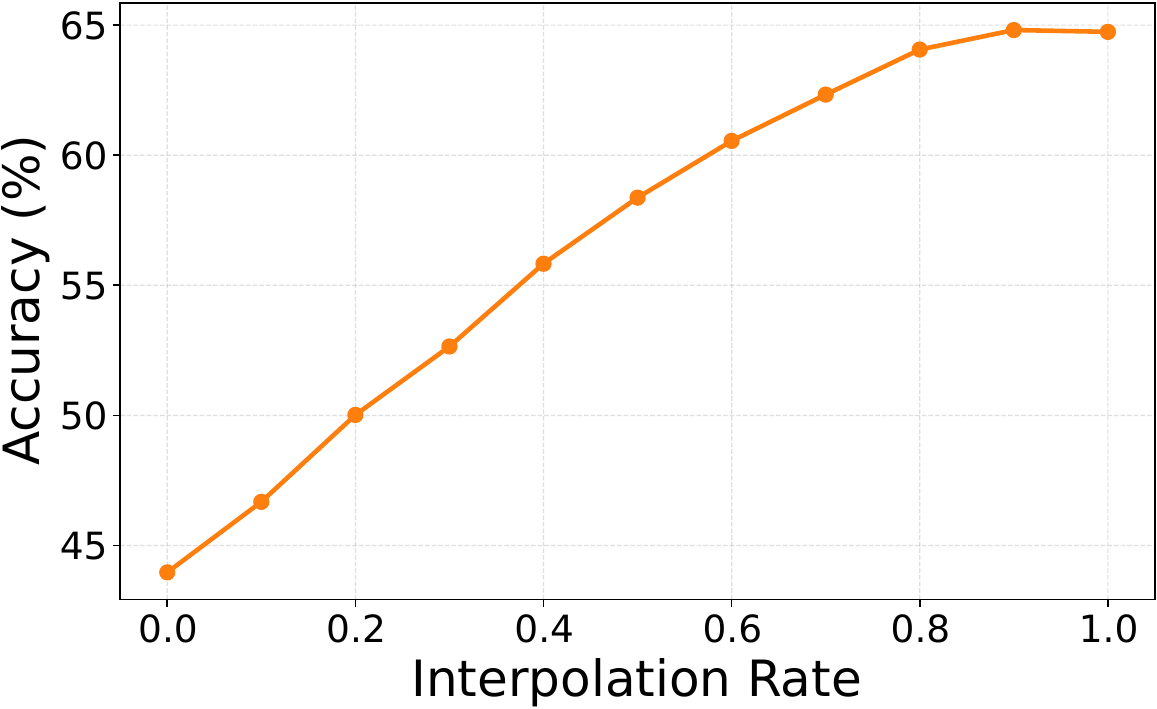}
    \caption{Training results after linearly interpolating high and low SNR data at different ratios.}
    \label{fig:group_c}
\end{subfigure}

\caption{Motivational analysis demonstrating that low-SNR degradation is a feature-collapse problem}
\label{fig:pre_exp}
\end{figure*}

\subsection{Motivation.}
While achieving robustness in low-SNR environments is a well-established challenge in the signal domain, how this problem manifests in large-scale MLLMs and how to best address it remains a question. Motivated by practices in the MLLM domain, a natural first attempt is to modify the composition of training data. We first explored this direction by pre-training several models with varying ratios of low-to-high SNR data.
As shown in Figure \ref{fig:group_a}, our experiments revealed a critical limitation of this data-centric approach. While increasing the volume of low-SNR training data provides a marginal benefit, it fails to yield any substantial improvement in low-SNR performance. This finding motivates our core thesis: the low-SNR problem in MLLMs must be addressed at the feature level, not merely at the data level.

Furthermore, our analyzes demonstrate that the low-SNR degradation originates from feature collapse. Specifically, as visualized in Figure \ref{fig:group_b}, low-SNR signals produce embeddings that are highly overlapped across classes, leading to semantic ambiguity and unstable multimodal alignment. Compellingly, directly manipulating these degraded embeddings—by linearly interpolating a low-SNR signal embedding towards its corresponding high-SNR version—resulted in a dramatic recovery of the LLM's generative performance, as evidenced in Figure \ref{fig:group_c}. This evidence strongly motivates a solution that targets the signal feature space. These findings form the cornerstone of our proposed MERLIN framework, which is designed to explicitly address this feature-level fragility.

\subsection{Model Architecture.} As illustrated in Figure \ref{fig:merlin}, the architecture of our baseline model consists of three integral comoponents: Signal Encoder, Signal Projector and Large Language Model (LLM).

To effectively obtain meaningful features from raw signal data, we employ the pretrained EMind, which has demonstrated state-of-the-art performance in analogous electromagnetic tasks. The encoder maps the signal into high-dimensional latent representations. Then the encoded features are then projected into the embedding space of the language model through a lightweight projector, which consists of two MLP layers with GELU activation. After projection, the signal embeddings are concatenated with the textual embeddings derived from the language tokenizer and jointly processed by a large language model (LLM) to generate language outpus.

\subsection{Two-Stage Training paradigm}
\cvprsubsubsection{Stage1: Foundational Pretraining}

To cultivate sophisticated reasoning abilities, we formulate the pretraining as a multi-task, instruction-following problem. Instead of using simple (signal, description) pairs, the model is trained to auto-regressively generate the answer text $A$ when presented with the signal $S$ and the question or instruction $Q$.

This instruction-tuning format trains the model to perform a wide range of tasks within a unified generative framework. The training objective is to minimize the standard next-token prediction loss:
$$\mathcal{L}_{pretrain}(\Theta) = -\sum_{i=1}^M\log P(a_i|a_{<i},Q,S;\Theta)$$
where $A = \{a_1, a_2, ..., a_M\}$ is the ground-truth answer of length $M$, while $P(a_i|a_{<i},Q,S; \Theta)$ represents the distribution for the response token $a_i$ given the preceding tokens $a_{<i}$, $Q$ and $S$. The entire set of model parameters, $\Theta = \{\theta_{enc}, \theta_{proj}, \theta_{llm}\}$, is jointly optimized to minimize this objective.
 are the sets of trainable parameters for the Signal Encoder, the Projector, and the LLM, respectively.





\cvprsubsubsection{Stage2: Low-SNR Robustness Enhancement}

To systematically correct the degraded signal embeddings, we introduce a comprehensive knowledge distillation framework. 
This framework comprises two models, a $Teacher$ and a $Student$, both initialized with the identical weights of the baseline model described in Section 3.1.
The $Teacher$, which operates as a static reference, remains frozen throughout this stage. The $Student$ is the target of our optimization. Its parameters are fully trainable and are updated during this stage.
Training is performed on a parallel dataset of tuples $(I_{high}, I_{low}, Q, A)$, where $I_{high}$ and $I_{low}$ respectively denote the high-SNR and low-SNR versions of the same underlying signal. In each step, $Teacher$ receives $(I_{high}, Q)$ and $Student$ receives $(I_{low}, Q)$. 

$Student$ is optimized using a composite objective that combines three distinct loss functions: the hard-label task loss $L_{task}$, the feature-level distillation loss $L_{feat}$, and the logit-level distillation loss $L_{logit}$.
These terms guide $Student$ toward both discriminative and semantically stable representations under noisy signal conditions. The final training objective combines the three components in a unified formulation:
$$\mathcal{L} = \mathcal{L}_{task} + \lambda_{logits}\mathcal{L}_{logits} + \lambda_{feat}\mathcal{L}_{feat}$$
where $\lambda_{logits}$ and $\lambda_{feat}$ are scalar hyperparameters that balance the contribution of each distillation term.

\textbf{Task Loss.} To ground $Student$ in the primary task, we compute a standar cross-entropy loss between its predictions and $A$, conditioned on the low-SNR input:
$$\mathcal{L}_{task} = -\sum_{i=1}^M\log P_{Student}(a_i|a_{<i}, Q, I_{low})$$

\textbf{Feature-level Distillation Loss.} To address the degradation of discriminative features under low-SNR conditions, we introduce a feature-level distillation mechanism that explicitly transfers the high-SNR structural characteristics of $Teacher$ to the low-SNR $Student$.
Given a pair of parallel signals, $Teacher$ projector outputs a high-quality embedding $f_{Teacher}$, while the student produces a noisy embedding $f_{Student}$.
However, direct alignment between $f_{Teacher}$ and $f_{Student}$ is often unstable due to noise contamination in $f_{Student}$.
To alleviate this issue, we employ a Denoising Subspace Module (DSM) that assumes the existence of an approximately orthogonal decomposition between signal and noise subspaces.
The DSM learns a projection matrix $P=UU^T$ spanning the signal subspace and projects the student embedding before computing the distillation loss:
$$\Phi(f_{Student}) = Pf_{Student}$$
$$L_{feat} = \Vert\Phi(f_{Student}) - f_{Student}\Vert_2^2$$

This denoising projection effectively filters out noise-dominated components, stabilizing the optimization process and “softening” the distillation.
Through this mechanism, the student learns to reconstruct high-SNR-like features even from degraded inputs, which fundamentally improves feature discriminability in low-SNR regimes.

\textbf{Logits-level Distillation Loss.}
To further ensure semantic consistency at the language modeling level, we apply a logit-level distillation loss that aligns the output distributions of the teacher and student LLMs.
Specifically, we minimize the KL divergence between their softened logits:
$$L_{logit} = T^2KL(Softmax(\frac{z_{Student}}{T}), Softmax(\frac{z_{Teacher}}{T}))$$
where $z_{Student}$ and $z_{Teacher}$ respectively denote $Teacher$ and $Student$ logits from the LLM head, and $T$ is the temperature coefficient.

\section{Experiment}
\label{sec:experiment}

\subsection{Experimental Setup}
\textbf{Implementation Details}
The architecture of our model consists of an EMind\cite{luo2025emind} signal encoder, a 2-layer MLP projector, and the Qwen3-4B-Instruct-2507 model as the LLM backbone. For both training stages, we use the AdamW optimizer and a cosine learning rate scheduler. Each stage is trained for a maximum of 8 epochs with a global batch size of 256 and an initial learning rate of 5e-5. The best checkpoint for each stage is selected via early stopping, based on the validation loss over a 10\% held-out portion of the training data. All experiments were conducted on a single server equipped with 8 NVIDIA A100 (80GB) GPUs.

\textbf{Dataset and Benchmark.} We employ a two-stage training process with final performance evaluated on the EM-Bench. The first pre-training stage utilizes the full EM-134K dataset. For the second stage, we construct a training set by generating low-SNR counterparts for high-SNR signals from EM-134K via Gaussian noise injection. This dataset of signal pairs is supplemented with a data replay mechanism from the original pre-training set to maintain general performance.

\textbf{Baselines.}
Our primary baseline involves feeding textualized signal data directly to a state-of-the-art LLM in a zero-shot setting. Raw IQ sequences are normalized and formatted into numerical strings embedded within task-specific prompts. This baseline is designed to quantify the signal processing capabilities inherent to the LLM itself, thereby isolating the performance gains provided by our multimodal architecture.

\begin{table*}[t]
\centering
\caption{\textbf{Main results on EM-Bench.} We compare MERLIN with leading proprietary and open-source LLMs. All baseline LLMs process EM signals in a textualized format. The best results are highlighted in \textbf{bold}. ''PE'' denotes Parameter Estimation tasks. Its sub-tasks denote \textbf{BW}: Bandwidth, \textbf{DC}: Duty Cycle, \textbf{NP}: Num. Pulses, \textbf{PRF}: Pulse Rep. Freq, \textbf{PW}: Pulse Width. MERLIN achieves state-of-the-art performance across both perception and reasoning tasks.}
\label{tab:main_results}
\resizebox{\textwidth}{!}{
\begin{tabular}{l|ccccccccccc|cccc}
\toprule
\multirow{3}{*}{\textbf{Model}} & \multicolumn{11}{c|}{\textbf{Perception Tasks}} & \multicolumn{4}{c}{\textbf{Reasoning Tasks}} \\
\cmidrule(lr){2-12} \cmidrule(lr){13-16}
 & \multirow{2}{*}{MOD} & \multicolumn{5}{c}{PE (Radar)} & \multirow{2}{*}{PI} & \multirow{2}{*}{CJR} & \multirow{2}{*}{RJR} & \multirow{2}{*}{SD} & \multirow{2}{*}{Avg.} & Anti-CJ & Anti-RJ & CLS & RJS \\
\cmidrule(lr){3-7}
\cmidrule(lr){13-16}
 &  & BW & DC & NP & PRF & PW &  &  &  &  &  & Rouge-L/BLEU & Rouge/BLEU & Rouge/BLEU & Rouge/BLEU\\
\midrule
GPT-5 & 28.00 & 12.67 & 0.00 & 52.67 & 14.00 & 6.67 & 30.00 & 19.67 & 26.00 & 42.33 & 23.20
       & 0.01 / 0.00 & 0.00 / 0.00 & 0.03 / 0.00 & 0.01 / 0.00  \\
Claude-4-Sonnet & 30.00 & 24.00 & 38.00 & 52.00 & 25.17 & 31.00 & 29.67 & 25.67 & 30.67 & 44.33 & 32.35
       & 0.11 / 0.00 & 0.10 / 0.00 & 0.10 / 0.00 & 0.12 / 0.01  \\
DeepSeek-v3.2-exp & 18.00 & 9.33 & 9.67 & 19.67 & 13.67 & 17.00 & 8.33 & 5.67 & 11.00 & 18.33 & 13.07
       & 0.05 / 0.00 & 0.06 / 0.00 & 0.09 / 0.00 & 0.09 / 0.00  \\
Qwen3-Next-80b-A3B & 17.00 & 7.33 & 14.67 & 29.33 & 15.33 & 16.33 & 10.00 & 9.33 & 13.67 & 24.00 & 15.70
       & 0.01 / 0.00 & 0.01 / 0.00 & 0.03 / 0.00 & 0.01 / 0.00  \\
Gemini-2.5-Pro & 24.00 & 8.74 & 50.67 & 52.33 & 17.20 & 15.63 & 29.67 & 24.67 & 31.00 & 45.25 & 29.92
       & 0.10 / 0.00 & 0.10 / 0.00 & 0.08 / 0.00 & 0.10 / 0.00  \\
\midrule
Qwen3-VL-4B-Instruct & \textbf{45.93} & \textbf{84.33} & 55.03 & 86.83 & 74.47 & 56.10 & \textbf{89.00} & 76.17 & 81.52 & \textbf{79.97} & 72.94
       & 0.43 / 0.14 & 0.31 / 0.09 & 0.31 / 0.06 & 0.28 / 0.10 \\
\midrule
\rowcolor{gray!15} \textbf{MERLIN (Ours)} & 44.97 & 82.13 & \textbf{75.30} & \textbf{92.40} & \textbf{82.77} & \textbf{72.53} & 87.37 & \textbf{86.00} & \textbf{83.04} & 76.20 & \textbf{78.27}
 & \textbf{0.45 / 0.15} & \textbf{0.29 / 0.08} & \textbf{0.40 / 0.08} & \textbf{0.42 / 0.22} \\
\bottomrule
\end{tabular}
}
\end{table*}

\subsection{Main Results}
To validate the efficacy of our proposed model, We evaluate MERLIN on EM-Bench against leading proprietary and open-source LLMs. Baselines process EM signals only as text, without a dedicated signal encoder. Results are summarized in Table \ref{tab:main_results}. In perception tasks, MERLIN exhibits a profound capability to accurately interpret raw signal characteristics. While baseline LLMs show some proficiency in simpler tasks, they falter on more nuanced parameter estimations. In contrast, MERLIN consistently delivers high accuracy, indicating its robust architecture is better suited for extracting detailed information from specialized electromagnetic data. This performance gap is even more pronounced in the reasoning tasks. MERLIN demonstrates a strong ability to analyze complex scenarios, such as identifying and responding to jamming signals. The baseline models, however, struggle to generate coherent or effective strategies, as reflected by their near-zero scores. In conclusion, the comparative analysis validates that MERLIN's specialized design provides a substantial leap in performance for both foundational perception and high-level reasoning within the electromagnetic domain, setting a new state-of-the-art.

\subsection{Ablation Study}

\begin{table}[t]
    \centering
    \setlength{\tabcolsep}{3.5pt}
    \caption{\textbf{Ablation of MERLIN's distillation components.} We show the impact of incrementally adding each component on model performance, starting from the pre-trained baseline.}
    \label{tab:ablation_distillation_compact}
    \begin{tabular}{l|ccc|cc}
        \toprule
        \multirow{2}{*}{\textbf{Method}} & \multicolumn{3}{c|}{\textbf{Components}} & \multicolumn{2}{c}{\textbf{Perf. (\%)}} \\
        \cmidrule(lr){2-4} \cmidrule(lr){5-6}
        & $\mathcal{L}_{MLP}$ & DSM & $\mathcal{L}_{Logits}$ & Low-SNR & Overall \\
        \midrule
        Stage-1 & $\times$ & $\times$ & $\times$ & 59.7 & 71.8 \\
        \midrule
        + Stage-2 & $\times$ & $\times$ & $\times$ & 62.9 & 77.6 \\
        + Feature KD   & \checkmark & $\times$ & $\times$ & 64.2 & 77.9 \\
        + DSM       & \checkmark & \checkmark & $\times$ & 64.4 & 78.0 \\
        \textbf{MERLIN} & \checkmark & \checkmark & \checkmark & \textbf{65.1} & \textbf{78.6} \\
        \bottomrule
    \end{tabular}
\end{table}

We conduct a detailed ablation study on EM-Bench to validate the contribution of each component in our framework. To facilitate a clear and quantitative analysis, this section reports the accuracy on the multiple-choice tasks of EM-Bench, with the full results presented in Table \ref{tab:ablation_distillation_compact}.

Our analysis begins with the model after its initial pre-training stage, which serves as our baseline (denoted as "Stage-1"). As expected, the model struggles with the specialized and noisy environment, establishing a clear need for further adaptation. A crucial finding is the effectiveness of a dedicated second training stage. Simply fine-tuning the pre-trained model on the target data results in a substantial leap in performance. This validates our two-stage approach, confirming that a specialized adaptation phase is essential for our model.

Building upon this strong fine-tuning baseline, we then demonstrate the superiority of our knowledge distillation strategy. By augmenting the conventional supervised objective with a MLP feature knowledge distillation loss, we observe a clear and consistent improvement. This result confirms our central hypothesis: for achieving noise robustness, it is effective to teach the model how to represent signals.
Finally, we show the value of our specific distillation enhancements. The integration of our DSM module provides a further improvement. The final inclusion of the logits loss completes the MERLIN framework and yields the highest performance, demonstrating the synergistic benefit of guiding the student model at both the feature and the output distribution levels. In conclusion, the ablation results systematically confirm that each component in our design is a deliberate  choice, working in concert to produce the final model's robustness.    
\section{Conclusion}
In this work, we establish a comprehensive foundation for MLLMs in the EM domain by systematically addressing three critical challenges. We introduce two key resources, the large-scale EM-134K dataset and the comprehensive EM-Bench benchmark, to resolve the foundational issues of data scarcity and the lack of standardized evaluation. Building upon this infrastructure, we propose MERLIN, a novel training framework designed to overcome the core challenge of performance degradation in low-SNR environments. Extensive experiments validate our integrated approach, demonstrating that MERLIN achieves state-of-the-art performance on EM-Bench and, crucially, exhibits exceptional robustness in noisy conditions where other methods falter. Together, these contributions provide a robust and validated pathway for future research into powerful MLLMs for the EM domain.

{
    \small
    \bibliographystyle{ieeenat_fullname}
    \bibliography{main}

\begin{thebibliography}{57}
\providecommand{\natexlab}[1]{#1}
\providecommand{\url}[1]{\texttt{#1}}
\expandafter\ifx\csname urlstyle\endcsname\relax
  \providecommand{\doi}[1]{doi: #1}\else
  \providecommand{\doi}{doi: \begingroup \urlstyle{rm}\Url}\fi

\bibitem[Achiam et~al.(2023)Achiam, Adler, Agarwal, Ahmad, Akkaya, Aleman, Almeida, Altenschmidt, Altman, Anadkat, et~al.]{achiam2023gpt}
Josh Achiam, Steven Adler, Sandhini Agarwal, Lama Ahmad, Ilge Akkaya, Florencia~Leoni Aleman, Diogo Almeida, Janko Altenschmidt, Sam Altman, Shyamal Anadkat, et~al.
\newblock Gpt-4 technical report.
\newblock \emph{arXiv preprint arXiv:2303.08774}, 2023.

\bibitem[Ahmed et~al.(2025)Ahmed, Islam, Datta, Kabir, Chowdhury, and Haque]{ahmed2025qwen}
Imtiaz Ahmed, Sadman Islam, Partha~Protim Datta, Imran Kabir, Naseef Ur~Rahman Chowdhury, and Ahshanul Haque.
\newblock Qwen 2.5: A comprehensive review of the leading resource-efficient llm with potentioal to surpass all competitors.
\newblock \emph{Authorea Preprints}, 2025.

\bibitem[Ahn et~al.(2025)Ahn, Kim, Park, Kim, and Lee]{ahn2025integrating}
Kyusu Ahn, Jinpyo Kim, Chanwoo Park, JiSoo Kim, and Jaejin Lee.
\newblock Integrating spatial and frequency information for under-display camera image restoration.
\newblock \emph{Pattern Analysis and Applications}, 28\penalty0 (4):\penalty0 184, 2025.

\bibitem[Alayrac et~al.(2022)Alayrac, Donahue, Luc, Miech, Barr, Hasson, Lenc, Mensch, Millican, Reynolds, et~al.]{alayrac2022flamingo}
Jean-Baptiste Alayrac, Jeff Donahue, Pauline Luc, Antoine Miech, Iain Barr, Yana Hasson, Karel Lenc, Arthur Mensch, Katherine Millican, Malcolm Reynolds, et~al.
\newblock Flamingo: a visual language model for few-shot learning.
\newblock \emph{Advances in neural information processing systems}, 35:\penalty0 23716--23736, 2022.

\bibitem[Anil et~al.(2023)Anil, Dai, Firat, Johnson, Lepikhin, Passos, Shakeri, Taropa, Bailey, Chen, et~al.]{anil2023palm}
Rohan Anil, Andrew~M Dai, Orhan Firat, Melvin Johnson, Dmitry Lepikhin, Alexandre Passos, Siamak Shakeri, Emanuel Taropa, Paige Bailey, Zhifeng Chen, et~al.
\newblock Palm 2 technical report.
\newblock \emph{arXiv preprint arXiv:2305.10403}, 2023.

\bibitem[Bai et~al.(2023)Bai, Bai, Yang, Wang, Tan, Wang, Lin, Zhou, and Zhou]{bai2023qwen}
Jinze Bai, Shuai Bai, Shusheng Yang, Shijie Wang, Sinan Tan, Peng Wang, Junyang Lin, Chang Zhou, and Jingren Zhou.
\newblock Qwen-vl: A frontier large vision-language model with versatile abilities.
\newblock \emph{arXiv preprint arXiv:2308.12966}, 1\penalty0 (2):\penalty0 3, 2023.

\bibitem[Basheer et~al.(2024)Basheer, Aldehim, Alluhaidan, and Sakri]{basheer2024improving}
Shakila Basheer, Ghadah Aldehim, Ala~Saleh Alluhaidan, and Sapiah Sakri.
\newblock Improving mental dysfunction detection from eeg signals: Self-contrastive learning and multitask learning with transformers.
\newblock \emph{Alexandria Engineering Journal}, 106:\penalty0 52--59, 2024.

\bibitem[Chen et~al.(2024{\natexlab{a}})Chen, Liu, Zhou, Liu, Tan, Li, Zhao, Qian, and Wei]{chen2024vall}
Sanyuan Chen, Shujie Liu, Long Zhou, Yanqing Liu, Xu Tan, Jinyu Li, Sheng Zhao, Yao Qian, and Furu Wei.
\newblock Vall-e 2: Neural codec language models are human parity zero-shot text to speech synthesizers.
\newblock \emph{arXiv preprint arXiv:2406.05370}, 2024{\natexlab{a}}.

\bibitem[Chen et~al.(2025)Chen, Zu, Feng, Yang, and Li]{chen2025radiollm}
Shuai Chen, Yong Zu, Zhixi Feng, Shuyuan Yang, and Mengchang Li.
\newblock Radiollm: Introducing large language model into cognitive radio via hybrid prompt and token reprogrammings.
\newblock \emph{arXiv preprint arXiv:2501.17888}, 2025.

\bibitem[Chen et~al.(2023)Chen, Hirschberg, and Tsao]{chen2023noise}
Yu-Wen Chen, Julia Hirschberg, and Yu Tsao.
\newblock Noise robust speech emotion recognition with signal-to-noise ratio adapting speech enhancement.
\newblock \emph{arXiv preprint arXiv:2309.01164}, 2023.

\bibitem[Chen et~al.(2024{\natexlab{b}})Chen, Wu, Wang, Su, Chen, Xing, Zhong, Zhang, Zhu, Lu, et~al.]{chen2024internvl}
Zhe Chen, Jiannan Wu, Wenhai Wang, Weijie Su, Guo Chen, Sen Xing, Muyan Zhong, Qinglong Zhang, Xizhou Zhu, Lewei Lu, et~al.
\newblock Internvl: Scaling up vision foundation models and aligning for generic visual-linguistic tasks.
\newblock In \emph{Proceedings of the IEEE/CVF conference on computer vision and pattern recognition}, pages 24185--24198, 2024{\natexlab{b}}.

\bibitem[Cheng et~al.(2025)Cheng, Zhang, Di, Niyato, and Song]{cheng2025large}
Lu Cheng, Hongliang Zhang, Boya Di, Dusit Niyato, and Lingyang Song.
\newblock Large language models empower multimodal integrated sensing and communication.
\newblock \emph{IEEE Communications Magazine}, 2025.

\bibitem[Comanici et~al.(2025)Comanici, Bieber, Schaekermann, Pasupat, Sachdeva, Dhillon, Blistein, Ram, Zhang, Rosen, et~al.]{comanici2025gemini}
Gheorghe Comanici, Eric Bieber, Mike Schaekermann, Ice Pasupat, Noveen Sachdeva, Inderjit Dhillon, Marcel Blistein, Ori Ram, Dan Zhang, Evan Rosen, et~al.
\newblock Gemini 2.5: Pushing the frontier with advanced reasoning, multimodality, long context, and next generation agentic capabilities.
\newblock \emph{arXiv preprint arXiv:2507.06261}, 2025.

\bibitem[Davaslioglu et~al.(2022)Davaslioglu, Bozta{\c{s}}, Ertem, Sagduyu, and Ayanoglu]{davaslioglu2022self}
Kemal Davaslioglu, Serdar Bozta{\c{s}}, Mehmet~Can Ertem, Yalin~E Sagduyu, and Ender Ayanoglu.
\newblock Self-supervised rf signal representation learning for nextg signal classification with deep learning.
\newblock \emph{IEEE Wireless Communications Letters}, 12\penalty0 (1):\penalty0 65--69, 2022.

\bibitem[Huang et~al.(2023)Huang, Pemasiri, Denman, Fookes, and Martin]{huang2023multi}
Zi Huang, Akila Pemasiri, Simon Denman, Clinton Fookes, and Terrence Martin.
\newblock Multi-task learning for radar signal characterisation.
\newblock In \emph{2023 IEEE International Conference on Acoustics, Speech, and Signal Processing Workshops (ICASSPW)}, pages 1--5. IEEE, 2023.

\bibitem[Im et~al.(2025)Im, Kim, and Kwon]{im2025cocl}
Hyeon-Jin Im, Jiye Kim, and Sunyoung Kwon.
\newblock Cocl: Eeg connectivity-guided contrastive learning for seizure detection.
\newblock \emph{ICT Express}, 2025.

\bibitem[Jagannath and Jagannath(2021)]{jagannath2021dataset}
Anu Jagannath and Jithin Jagannath.
\newblock Dataset for modulation classification and signal type classification for multi-task and single task learning.
\newblock \emph{Computer Networks}, 199:\penalty0 108441, 2021.

\bibitem[Jebur et~al.(2024)Jebur, Zabil, Hammood, and Cheng]{jebur2024comprehensive}
Rusul~Sabah Jebur, Mohd Hazli Bin~Mohamed Zabil, Dalal~Adulmohsin Hammood, and Lim~Kok Cheng.
\newblock A comprehensive review of image denoising in deep learning.
\newblock \emph{Multimedia Tools and Applications}, 83\penalty0 (20):\penalty0 58181--58199, 2024.

\bibitem[Jiang et~al.(2023)Jiang, Wang, Xu, Sun, Gonzalez, Chen, Wu, Xiang, and Ren]{jiang2023radiation}
Zhuoran Jiang, Siqi Wang, Yifei Xu, Leshan Sun, Gilberto Gonzalez, Yong Chen, Q~Jackie Wu, Liangzhong Xiang, and Lei Ren.
\newblock Radiation-induced acoustic signal denoising using a supervised deep learning framework for imaging and therapy monitoring.
\newblock \emph{Physics in Medicine \& Biology}, 68\penalty0 (23):\penalty0 235010, 2023.

\bibitem[Jovhari et~al.(2025)Jovhari, Sedaghat, Shah-Hosseini, Mohammadi, and Hasanlou]{jovhari2025noise}
Negar Jovhari, Amin Sedaghat, Reza Shah-Hosseini, Nazila Mohammadi, and Mahdi Hasanlou.
\newblock Noise-robust multimodal remote sensing image matching via geometric analysis of embedded pre-trained manifolds.
\newblock \emph{Neurocomputing}, 638:\penalty0 130150, 2025.

\bibitem[Kang et~al.(2024)Kang, Han, and Hong]{kang2024sensing}
Seonghyeon Kang, Kawon Han, and Songcheol Hong.
\newblock Sensing-aided distortion estimation for ofdm radar with nonlinear transmitter.
\newblock \emph{IEEE Transactions on Radar Systems}, 2024.

\bibitem[Khudjaev et~al.(2024)Khudjaev, Tsoy, A~Sharif, Myrzabekov, Kim, and Lee]{khudjaev2024dformer}
Nodirkhuja Khudjaev, Roman Tsoy, SM A~Sharif, Azamat Myrzabekov, Seongwan Kim, and Jaeho Lee.
\newblock Dformer: Learning efficient image restoration with perceptual guidance.
\newblock In \emph{Proceedings of the IEEE/CVF Conference on Computer Vision and Pattern Recognition}, pages 6363--6372, 2024.

\bibitem[Kim and Chung(2024)]{kim2024vit}
Gyu-Il Kim and Kyungyong Chung.
\newblock Vit-based multi-scale classification using digital signal processing and image transformation.
\newblock \emph{IEEE Access}, 12:\penalty0 58625--58638, 2024.

\bibitem[Li et~al.(2024)Li, Zhang, Zhang, Zhang, Li, Li, Ma, and Li]{li2024llava}
Feng Li, Renrui Zhang, Hao Zhang, Yuanhan Zhang, Bo Li, Wei Li, Zejun Ma, and Chunyuan Li.
\newblock Llava-next-interleave: Tackling multi-image, video, and 3d in large multimodal models.
\newblock \emph{arXiv preprint arXiv:2407.07895}, 2024.

\bibitem[Li et~al.(2023{\natexlab{a}})Li, Li, Savarese, and Hoi]{li2023blip}
Junnan Li, Dongxu Li, Silvio Savarese, and Steven Hoi.
\newblock Blip-2: Bootstrapping language-image pre-training with frozen image encoders and large language models.
\newblock In \emph{International conference on machine learning}, pages 19730--19742. PMLR, 2023{\natexlab{a}}.

\bibitem[Li et~al.(2023{\natexlab{b}})Li, He, Wang, Li, Wang, Luo, Wang, Wang, and Qiao]{li2023videochat}
KunChang Li, Yinan He, Yi Wang, Yizhuo Li, Wenhai Wang, Ping Luo, Yali Wang, Limin Wang, and Yu Qiao.
\newblock Videochat: Chat-centric video understanding.
\newblock \emph{arXiv preprint arXiv:2305.06355}, 2023{\natexlab{b}}.

\bibitem[Li and Zhu(2024)]{li2024noisy}
Mengting Li and Chuang Zhu.
\newblock Noisy label processing for classification: A survey.
\newblock \emph{arXiv preprint arXiv:2404.04159}, 2024.

\bibitem[Lin et~al.(2024)Lin, Ye, Zhu, Cui, Ning, Jin, and Yuan]{lin2024video}
Bin Lin, Yang Ye, Bin Zhu, Jiaxi Cui, Munan Ning, Peng Jin, and Li Yuan.
\newblock Video-llava: Learning united visual representation by alignment before projection.
\newblock In \emph{Proceedings of the 2024 Conference on Empirical Methods in Natural Language Processing}, pages 5971--5984, 2024.

\bibitem[Liu et~al.(2024)Liu, Zhao, Kuang, Kang, Jiang, Sun, and Wu]{liu2024evolving}
Chengyuan Liu, Fubang Zhao, Kun Kuang, Yangyang Kang, Zhuoren Jiang, Changlong Sun, and Fei Wu.
\newblock Evolving knowledge distillation with large language models and active learning.
\newblock In \emph{Proceedings of the 2024 Joint International Conference on Computational Linguistics, Language Resources and Evaluation (LREC-COLING 2024)}, pages 6717--6731, 2024.

\bibitem[Liu et~al.(2025)Liu, Wang, Mao, Niyato, Wang, and Gui]{liu2025spectrumllm}
Chao Liu, Yu Wang, Shiwen Mao, Dusit Niyato, Xianbin Wang, and Guan Gui.
\newblock Spectrumllm: Large language models for next-generation spectrum prediction.
\newblock \emph{IEEE Wireless Communications}, 2025.

\bibitem[Liu et~al.(2023)Liu, Li, Wu, and Lee]{liu2023visual}
Haotian Liu, Chunyuan Li, Qingyang Wu, and Yong~Jae Lee.
\newblock Visual instruction tuning.
\newblock \emph{Advances in neural information processing systems}, 36:\penalty0 34892--34916, 2023.

\bibitem[Luo et~al.(2025)Luo, Gui, Liu, Zhang, Zhang, Wang, Guo, Ma, Liu, He, et~al.]{luo2025emind}
Luqing Luo, Wenjin Gui, Yunfei Liu, Ziyue Zhang, Yunxi Zhang, Fengxiang Wang, Zonghao Guo, Zizhi Ma, Xinzhu Liu, Hanxiang He, et~al.
\newblock Emind: A foundation model for multi-task electromagnetic signals understanding.
\newblock \emph{arXiv preprint arXiv:2508.18785}, 2025.

\bibitem[Lyu et~al.(2024)Lyu, Yan, Wang, Shi, Yin, Ren, Chen, de~Rijke, and Ren]{lyu2024knowtuning}
Yougang Lyu, Lingyong Yan, Shuaiqiang Wang, Haibo Shi, Dawei Yin, Pengjie Ren, Zhumin Chen, Maarten de Rijke, and Zhaochun Ren.
\newblock Knowtuning: Knowledge-aware fine-tuning for large language models.
\newblock \emph{arXiv preprint arXiv:2402.11176}, 2024.

\bibitem[Makwana et~al.(2024)Makwana, Deshmukh, Susladkar, Mittal, et~al.]{makwana2024livenet}
Dhruv Makwana, Gayatri Deshmukh, Onkar Susladkar, Sparsh Mittal, et~al.
\newblock Livenet: A novel network for real-world low-light image denoising and enhancement.
\newblock In \emph{Proceedings of the IEEE/CVF Winter Conference on Applications of Computer Vision}, pages 5856--5865, 2024.

\bibitem[Oh and Bui(2025)]{oh2025multi}
YongKyung Oh and Alex Bui.
\newblock Multi-view contrastive learning for robust domain adaptation in medical time series analysis.
\newblock \emph{arXiv preprint arXiv:2506.22393}, 2025.

\bibitem[O'shea and West(2016)]{o2016radio}
Timothy~J O'shea and Nathan West.
\newblock Radio machine learning dataset generation with gnu radio.
\newblock In \emph{Proceedings of the GNU radio conference}, 2016.

\bibitem[O’Shea et~al.(2018)O’Shea, Roy, and Clancy]{o2018over}
Timothy~James O’Shea, Tamoghna Roy, and T~Charles Clancy.
\newblock Over-the-air deep learning based radio signal classification.
\newblock \emph{IEEE Journal of Selected Topics in Signal Processing}, 12\penalty0 (1):\penalty0 168--179, 2018.

\bibitem[Peng and Zhang(2025)]{peng2025enhancing}
Tianyu Peng and Jiajun Zhang.
\newblock Enhancing knowledge distillation of large language models through efficient multi-modal distribution alignment.
\newblock In \emph{Proceedings of the 31st International Conference on Computational Linguistics}, pages 2478--2496, 2025.

\bibitem[Quan et~al.(2025)Quan, Cheng, Yang, Zhao, Chen, and Luo]{quan2025signal}
Wei Quan, Wenjing Cheng, Yike Yang, Haiquan Zhao, Zhaoyu Chen, and Yunfan Luo.
\newblock A signal fingerprint feature extraction method based on decomposition and fusion for radar emitter individual identification.
\newblock \emph{Digital Signal Processing}, page 105257, 2025.

\bibitem[Rubenstein et~al.(2023)Rubenstein, Asawaroengchai, Nguyen, Bapna, Borsos, Quitry, Chen, Badawy, Han, Kharitonov, et~al.]{rubenstein2023audiopalm}
Paul~K Rubenstein, Chulayuth Asawaroengchai, Duc~Dung Nguyen, Ankur Bapna, Zal{\'a}n Borsos, F{\'e}lix de~Chaumont Quitry, Peter Chen, Dalia~El Badawy, Wei Han, Eugene Kharitonov, et~al.
\newblock Audiopalm: A large language model that can speak and listen.
\newblock \emph{arXiv preprint arXiv:2306.12925}, 2023.

\bibitem[Shao et~al.(2024)Shao, Tong, Wu, Guo, Li, Lin, and Zhang]{shao2024wirelessllm}
Jiawei Shao, Jingwen Tong, Qiong Wu, Wei Guo, Zijian Li, Zehong Lin, and Jun Zhang.
\newblock Wirelessllm: Empowering large language models towards wireless intelligence.
\newblock \emph{arXiv preprint arXiv:2405.17053}, 2024.

\bibitem[Shi et~al.(2024)Shi, Zhang, Guo, Yang, Xu, and Wu]{shi2024survey}
Jialin Shi, Kailai Zhang, Chenyi Guo, Youquan Yang, Yali Xu, and Ji Wu.
\newblock A survey of label-noise deep learning for medical image analysis.
\newblock \emph{Medical image analysis}, 95:\penalty0 103166, 2024.

\bibitem[Taassori(2025)]{taassori2025robustdeit}
Mehdi Taassori.
\newblock Robustdeit: Noise-robust vision transformers for medical image classification.
\newblock \emph{Image Analysis and Stereology}, 44\penalty0 (2):\penalty0 111--129, 2025.

\bibitem[Tang et~al.(2023)Tang, Yu, Sun, Chen, Tan, Li, Lu, Ma, and Zhang]{tang2023salmonn}
Changli Tang, Wenyi Yu, Guangzhi Sun, Xianzhao Chen, Tian Tan, Wei Li, Lu Lu, Zejun Ma, and Chao Zhang.
\newblock Salmonn: Towards generic hearing abilities for large language models.
\newblock \emph{arXiv preprint arXiv:2310.13289}, 2023.

\bibitem[Tao et~al.(2025)Tao, Wang, Zhang, Wang, and Gu]{tao2025nevlp}
Yiyi Tao, Zhuoyue Wang, Hang Zhang, Lun Wang, and Jianxin Gu.
\newblock Nevlp: Noise-robust framework for efficient vision-language pre-training.
\newblock In \emph{International Conference on Intelligent Computing}, pages 74--85. Springer, 2025.

\bibitem[Team et~al.(2023)Team, Anil, Borgeaud, Alayrac, Yu, Soricut, Schalkwyk, Dai, Hauth, Millican, et~al.]{team2023gemini}
Gemini Team, Rohan Anil, Sebastian Borgeaud, Jean-Baptiste Alayrac, Jiahui Yu, Radu Soricut, Johan Schalkwyk, Andrew~M Dai, Anja Hauth, Katie Millican, et~al.
\newblock Gemini: a family of highly capable multimodal models.
\newblock \emph{arXiv preprint arXiv:2312.11805}, 2023.

\bibitem[Tekb{\i}y{\i}k et~al.(2019)Tekb{\i}y{\i}k, Ke{\c{c}}eci, Ekti, G{\"o}r{\c{c}}in, and Kurt]{tekbiyik2019hisarmod}
K Tekb{\i}y{\i}k, C Ke{\c{c}}eci, AR Ekti, A G{\"o}r{\c{c}}in, and G Kurt.
\newblock Hisarmod: A new challenging modulated signals dataset.
\newblock \emph{IEEE Dataport}, 2019.

\bibitem[Wang et~al.(2024{\natexlab{a}})Wang, Liu, He, Xu, Ma, and Li]{wang2024eegpt}
Guangyu Wang, Wenchao Liu, Yuhong He, Cong Xu, Lin Ma, and Haifeng Li.
\newblock Eegpt: Pretrained transformer for universal and reliable representation of eeg signals.
\newblock \emph{Advances in Neural Information Processing Systems}, 37:\penalty0 39249--39280, 2024{\natexlab{a}}.

\bibitem[Wang et~al.(2024{\natexlab{b}})Wang, Bai, Tan, Wang, Fan, Bai, Chen, Liu, Wang, Ge, et~al.]{wang2024qwen2}
Peng Wang, Shuai Bai, Sinan Tan, Shijie Wang, Zhihao Fan, Jinze Bai, Keqin Chen, Xuejing Liu, Jialin Wang, Wenbin Ge, et~al.
\newblock Qwen2-vl: Enhancing vision-language model's perception of the world at any resolution.
\newblock \emph{arXiv preprint arXiv:2409.12191}, 2024{\natexlab{b}}.

\bibitem[Weng et~al.(2025)Weng, Gu, Guo, Ma, Yang, Liu, and Chen]{weng2025self}
Weining Weng, Yang Gu, Shuai Guo, Yuan Ma, Zhaohua Yang, Yuchen Liu, and Yiqiang Chen.
\newblock Self-supervised learning for electroencephalogram: A systematic survey.
\newblock \emph{ACM Computing Surveys}, 57\penalty0 (12):\penalty0 1--38, 2025.

\bibitem[Xu et~al.(2022)Xu, Wang, Fu, and Jia]{xu2022snr}
Xiaogang Xu, Ruixing Wang, Chi-Wing Fu, and Jiaya Jia.
\newblock Snr-aware low-light image enhancement.
\newblock In \emph{Proceedings of the IEEE/CVF conference on computer vision and pattern recognition}, pages 17714--17724, 2022.

\bibitem[Yan et~al.(2021)Yan, Zhang, Abbeel, and Srinivas]{yan2021videogpt}
Wilson Yan, Yunzhi Zhang, Pieter Abbeel, and Aravind Srinivas.
\newblock Videogpt: Video generation using vq-vae and transformers.
\newblock \emph{arXiv preprint arXiv:2104.10157}, 2021.

\bibitem[Zhang et~al.(2023{\natexlab{a}})Zhang, Li, Zhang, Zhan, Wang, Zhou, and Qiu]{zhang2023speechgpt}
Dong Zhang, Shimin Li, Xin Zhang, Jun Zhan, Pengyu Wang, Yaqian Zhou, and Xipeng Qiu.
\newblock Speechgpt: Empowering large language models with intrinsic cross-modal conversational abilities.
\newblock \emph{arXiv preprint arXiv:2305.11000}, 2023{\natexlab{a}}.

\bibitem[Zhang et~al.(2023{\natexlab{b}})Zhang, Li, and Bing]{zhang2023video}
Hang Zhang, Xin Li, and Lidong Bing.
\newblock Video-llama: An instruction-tuned audio-visual language model for video understanding.
\newblock \emph{arXiv preprint arXiv:2306.02858}, 2023{\natexlab{b}}.

\bibitem[Zhang et~al.(2024)Zhang, Yan, Liu, Chen, Lau, Yu, and Shao]{zhang2024adaptive}
Jingming Zhang, Yaxi Yan, Shuaiqi Liu, Xingwei Chen, Alan Pak~Tao Lau, Changyuan Yu, and Liyang Shao.
\newblock Adaptive block-matching and 3d denoising for $\phi$-otdr under ultra-low snr conditions.
\newblock \emph{Journal of Lightwave Technology}, 42\penalty0 (13):\penalty0 4698--4705, 2024.

\bibitem[Zhou et~al.(2024)Zhou, Chen, Pan, Shi, and Yang]{zhou2024adapt}
Shihao Zhou, Duosheng Chen, Jinshan Pan, Jinglei Shi, and Jufeng Yang.
\newblock Adapt or perish: Adaptive sparse transformer with attentive feature refinement for image restoration.
\newblock In \emph{Proceedings of the IEEE/CVF conference on computer vision and pattern recognition}, pages 2952--2963, 2024.

\bibitem[Zhu et~al.(2023)Zhu, Chen, Shen, Li, and Elhoseiny]{zhu2023minigpt}
Deyao Zhu, Jun Chen, Xiaoqian Shen, Xiang Li, and Mohamed Elhoseiny.
\newblock Minigpt-4: Enhancing vision-language understanding with advanced large language models.
\newblock \emph{arXiv preprint arXiv:2304.10592}, 2023.

\end{thebibliography}
}

\clearpage

\appendix
\clearpage  
\onecolumn  
\section{Appendix}
\subsection{Overview of Appendix}
\label{Appendix Benchmark}
This appendix provides supplementary details for our pre-trained baseline, dataset, and benchmark. Due to space constraints in the main paper, additional technical specifics can only be accommodated here. 

The organization is as follows:

\begin{itemize}
    \item Sec. A.2: More implent details of baseline model.
    \item Sec. A.3: Datasheets for benchmark.
    \item Sec. A.4: Visualizations of samples and challenging cases.
    \item Sec. A.5: SNR Robustness: Pretrain Model vs. MERLIN.
    \item Sec. A.6: Prompts on comparative Experiments Across Different General Models
\end{itemize}
\subsection{Implementation Details of Baseline Model}
\label{sec:details_model}
We utilized EM-100K as the supervised fine-tuning (SFT) dataset. The pre-trained model is derived from Qwen3-8B, to which we first integrated a signal encoder (SIT) that embeds IQ signal data into feature vectors. These feature vectors serve as a second modality input alongside text into the LLM. 

During two-stage training, we perform full fine-tuning in the first stage to align signal and text semantics. In the second stage, we apply knowledge distillation fine-tuning to enhance the model’s robustness under low-SNR conditions, where the LLM parameters are frozen and only the signal encoder and projection layers are trained.

\begin{table}[ht] 
\centering
\caption{\textbf{Training configuration details for MERLIN framework.} Hyperparameters used during the two-stage training. SIT = Signal Integration Transformer encoder.}
\label{tab:training_config}
\begin{tabular}{ll}
\toprule
\textbf{Configuration} & \textbf{Parameter Value} \\ \midrule
Dataset (Stage 1) & EM-100K (100K instruction pairs) \\
Dataset (Stage 2) & Synthetic low-SNR variants (SNR = $-$20dB to 20dB) \\
Batch Size & 256 \\
LR: Signal Encoder (SIT) & $5.0 \times 10^{-5}$ (cosine decay) \\
LR: Projector, LLM & $5.0 \times 10^{-5}$ \\
Optimizer & AdamW ($\beta_1=0.9, \beta_2=0.999, \epsilon=1\text{e-}8$) \\
Training Epochs & Stage 1: 8 epochs, Stage 2: 8 epochs \\
Hardware & 8 $\times$ NVIDIA A100 80GB \\ \bottomrule
\end{tabular}
\end{table}

\subsection{Datasheets for benchmark}
\begin{table*}[!ht]
\centering
\renewcommand{\arraystretch}{1.25} 
\caption{\textbf{Overview of the EM-Bench evaluation framework.} The benchmark features a three-level hierarchy spanning Perception and Reasoning, consisting of 14 distinct L3-Tasks. Each sub-task contains 300 samples, totaling 4,200 VQA pairs. The table lists the specific abbreviations, data quantities, and answer types (Single Choice vs. Open-ended Strategy) for each task.}
\resizebox{\textwidth}{!}{
\begin{tabular}{c|c|ccccc}
\toprule
\rowcolor[HTML]{EFEFEF} 
\textbf{L1-Task} & \textbf{L2-Task} & \textbf{L3-Task} & \textbf{Abbr.} & \textbf{Annotation Format} & \textbf{Number of Samples} & \textbf{Answer Type} \\ \hline
\multirow{10}{*}{Perception} & \multirow{7}{*}{Signal Characterization} & Modulation Classification & MOD & VQA & 300 & Single Choice(A/B/C/D/E) \\
 &  & Duty Cycle Estimation & PE.DC & VQA & 300 & Single Choice(A/B/C/D/E) \\ 
 &  & Pulse Repetition Frequency Estimation & PE.PRF & VQA & 300 & Single Choice(A/B/C/D/E) \\ 
 &  & Bandwidth Estimation & PE.BW & VQA & 300 & Single Choice(A/B/C/D/E) \\ 
 &  & Pulse Width Estimation & PE.PW & VQA & 300 & Single Choice(A/B/C/D/E) \\ 
 &  & Pulse Number Estimation & PE.NoP & VQA & 300 & Single Choice(A/B/C/D/E)\\ 
 &  & Protocol Identification & PI & VQA & 300 & Single Choice(A/B/C/D/E) \\ \cline{2-7}
 & \multirow{2}{*}{Jamming Identification} & Radar Jamming Judgement & RJR & VQA & 300 & Single Choice(A/B/C/D/E) \\
 &  & Communication Jamming Judgement & CJR & VQA & 300 & Single Choice(A/B/C/D/E) \\ \cline{2-7}
 & Segment Detection & Jamming Segment Detection & SD & VQA & 300 & Single Choice(A/B/C/D/E) \\ \hline
\multirow{4}{*}{Reasoning} & \multirow{4}{*}{Strategy Generation} & Anti-Communication Jamming Strategy & Anti-CJ & VQA & 300 & Open-end Strategy \\ \cline{3-7}
 &  & Anti-Radar Jamming Strategy & Anti-RJ & VQA & 300 & Open-end Strategy \\ \cline{3-7}
 &  & Communication Jamming Strategy & CJS & VQA & 300 & Open-end Strategy \\ \cline{3-7}
 &  & Radar Jamming Strategy & RJS & VQA & 300 & Open-end Strategy \\ \hline
\end{tabular}
}
\label{tab:em-bench-appendix}
\end{table*}

\subsection{Detailed Performance Analysis across SNR Levels}
\label{subsec:snr_analysis}

\begin{figure*}[t!]
    \centering
    \begin{subfigure}{0.19\textwidth}
        \includegraphics[width=\linewidth]{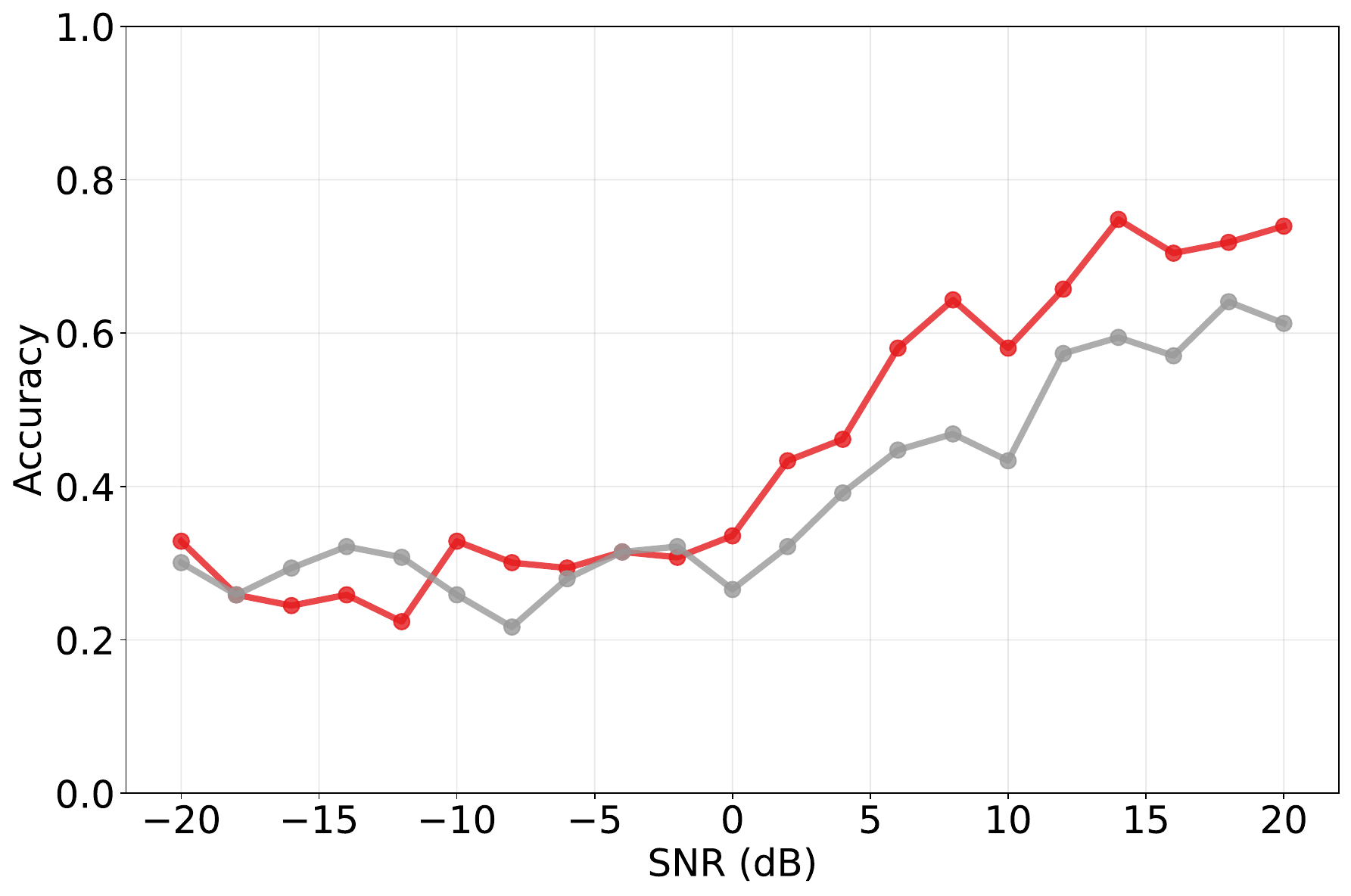} 
        \caption{Modulation}
    \end{subfigure}
    \hfill
    \begin{subfigure}{0.19\textwidth}
        \includegraphics[width=\linewidth]{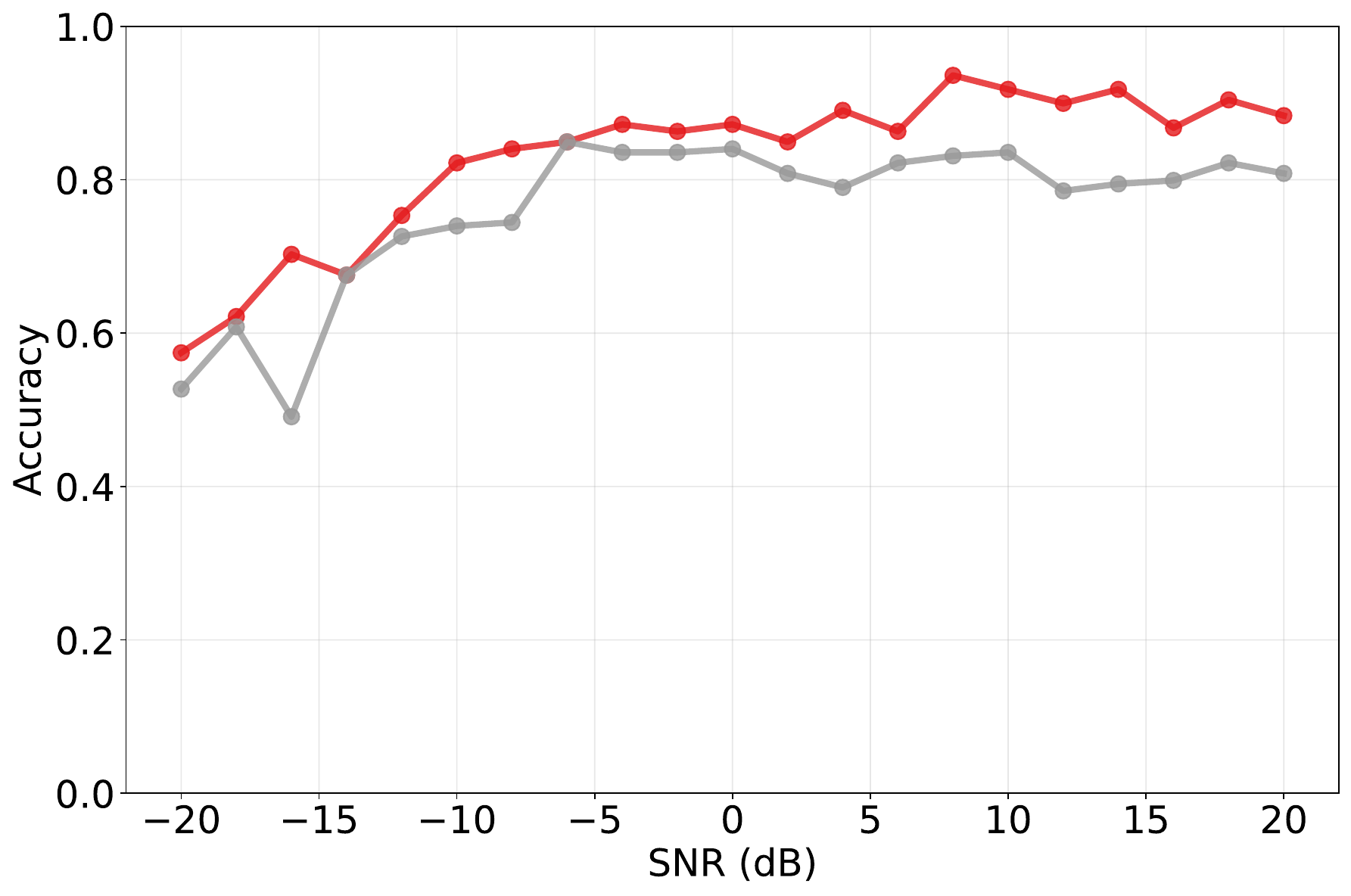}
        \caption{Bandwidth}
    \end{subfigure}
    \hfill
    \begin{subfigure}{0.19\textwidth}
        \includegraphics[width=\linewidth]{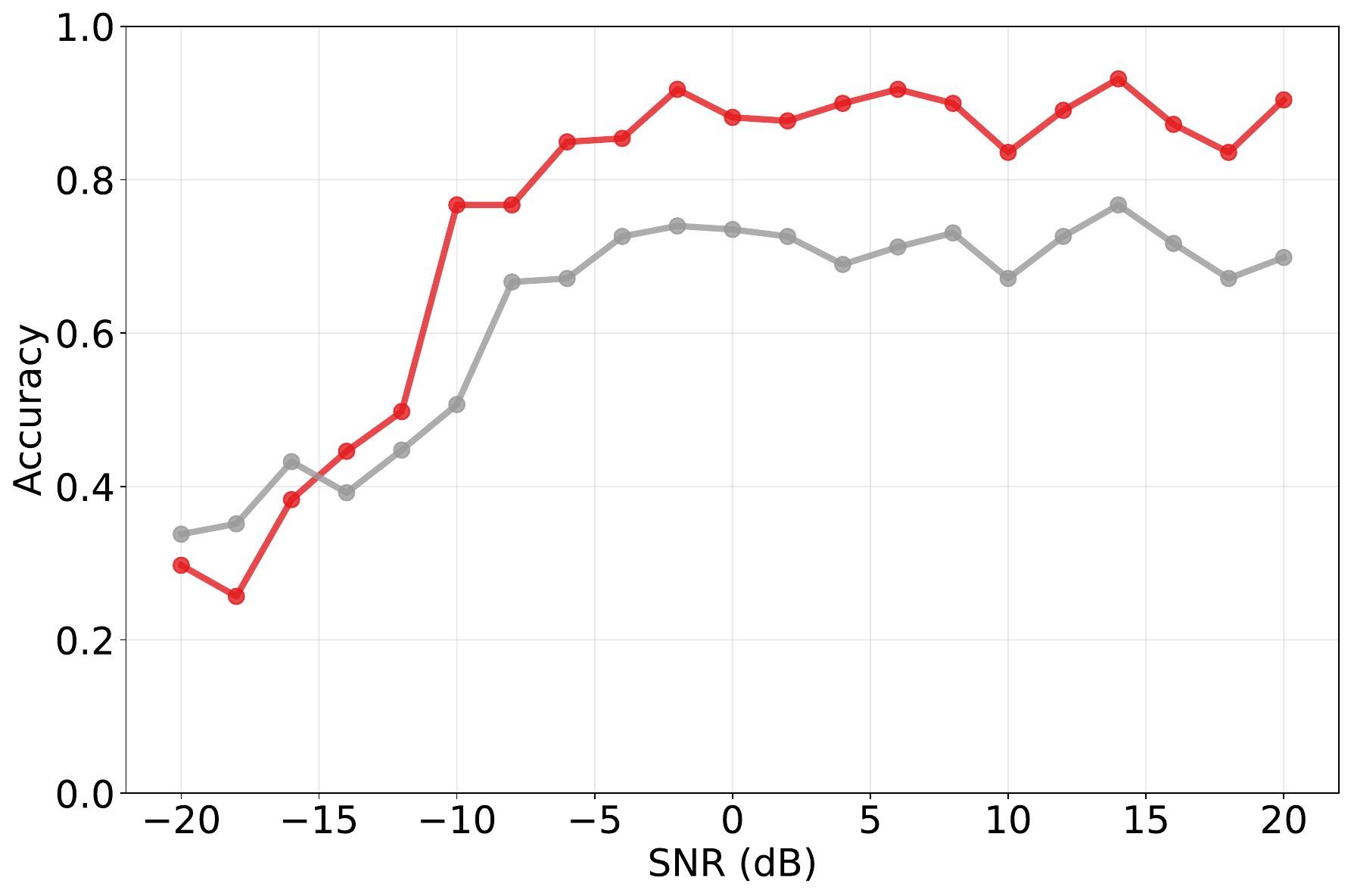}
        \caption{Duty Cycle}
    \end{subfigure}
    \hfill
    \begin{subfigure}{0.19\textwidth}
        \includegraphics[width=\linewidth]{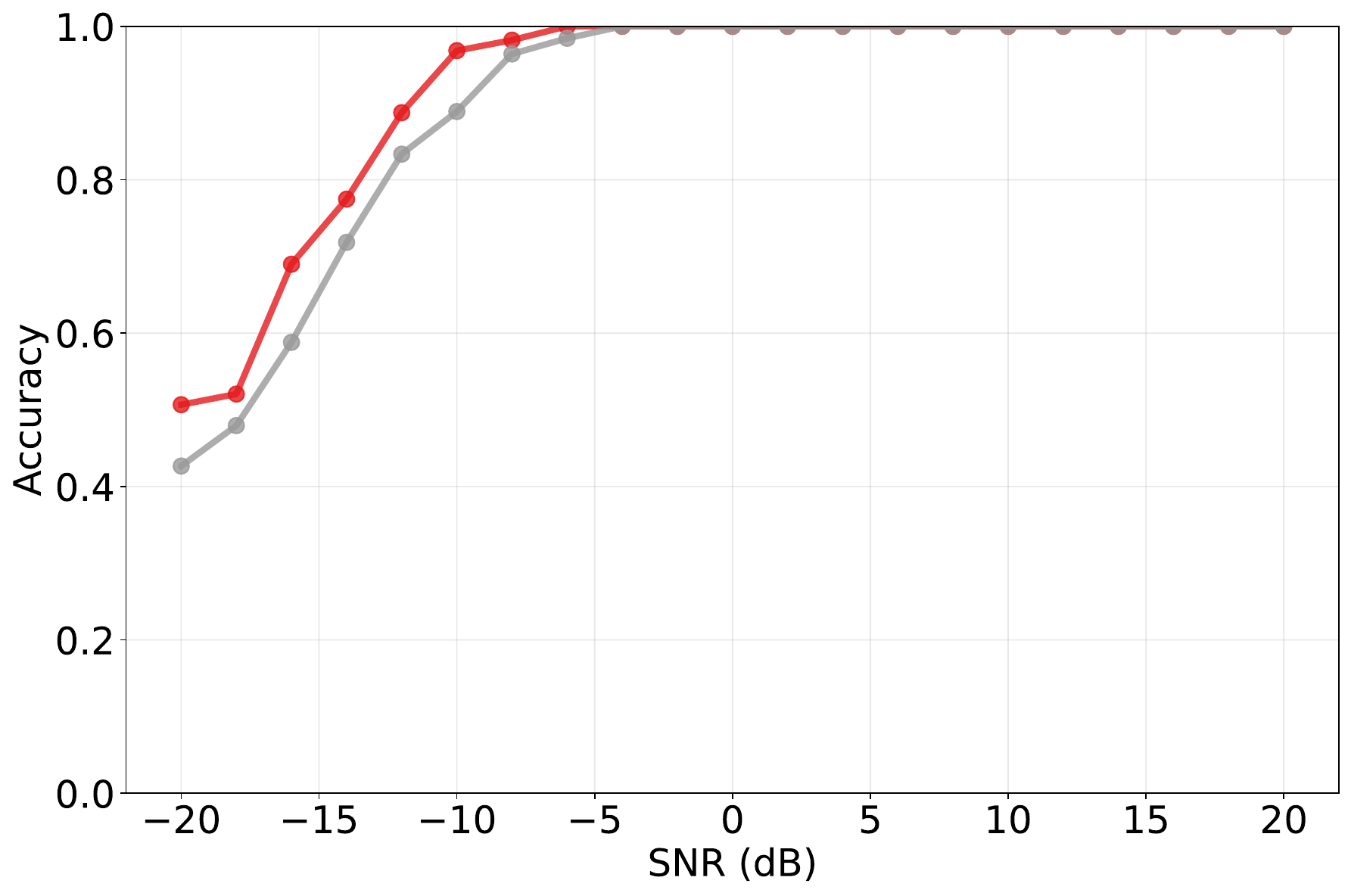}
        \caption{Num. Pulses}
    \end{subfigure}
    \hfill
    \begin{subfigure}{0.19\textwidth}
        \includegraphics[width=\linewidth]{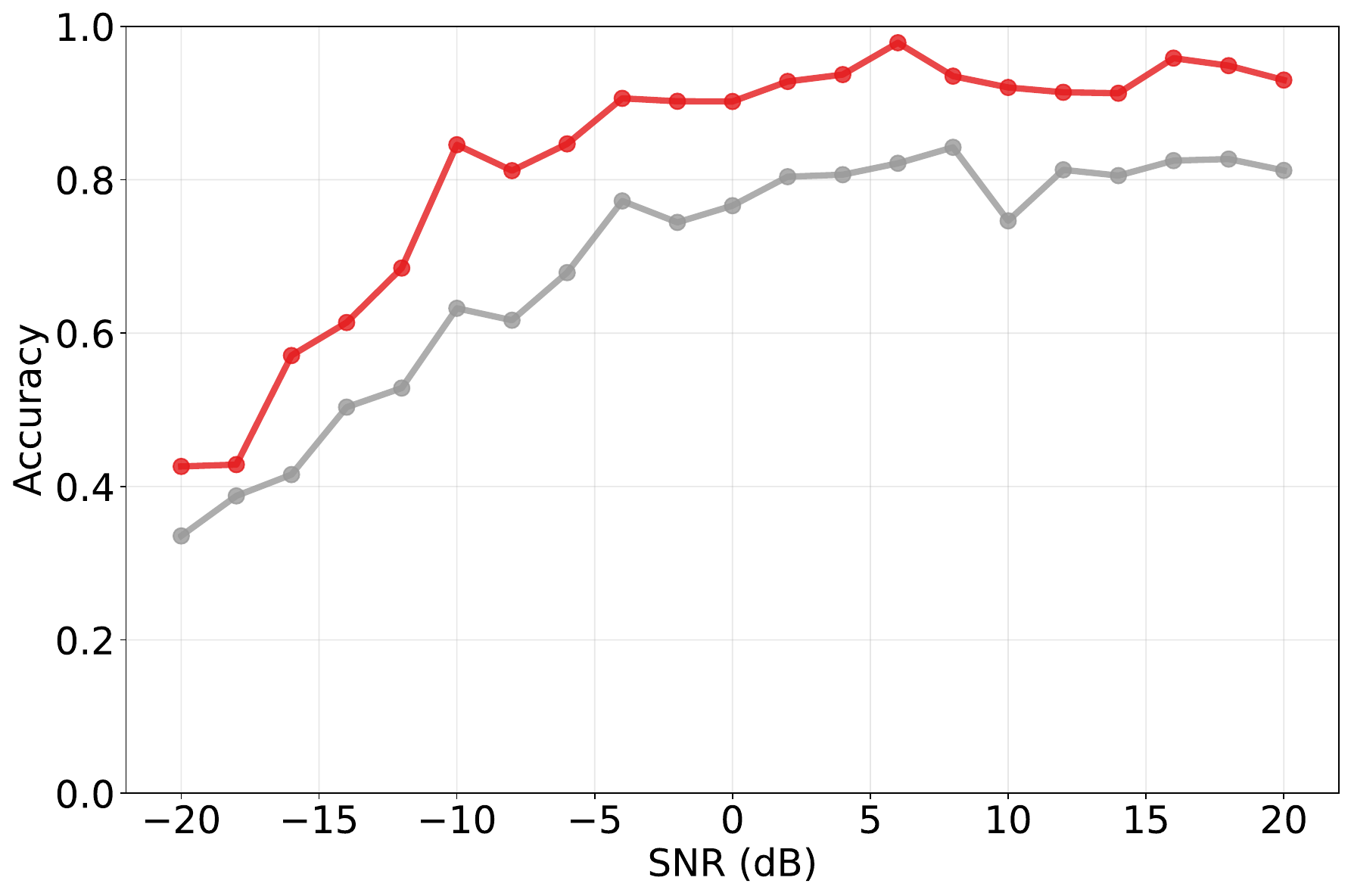}
        \caption{PRF Est.}
    \end{subfigure}
    
    \vspace{1mm}
    
    \begin{subfigure}{0.19\textwidth}
        \includegraphics[width=\linewidth]{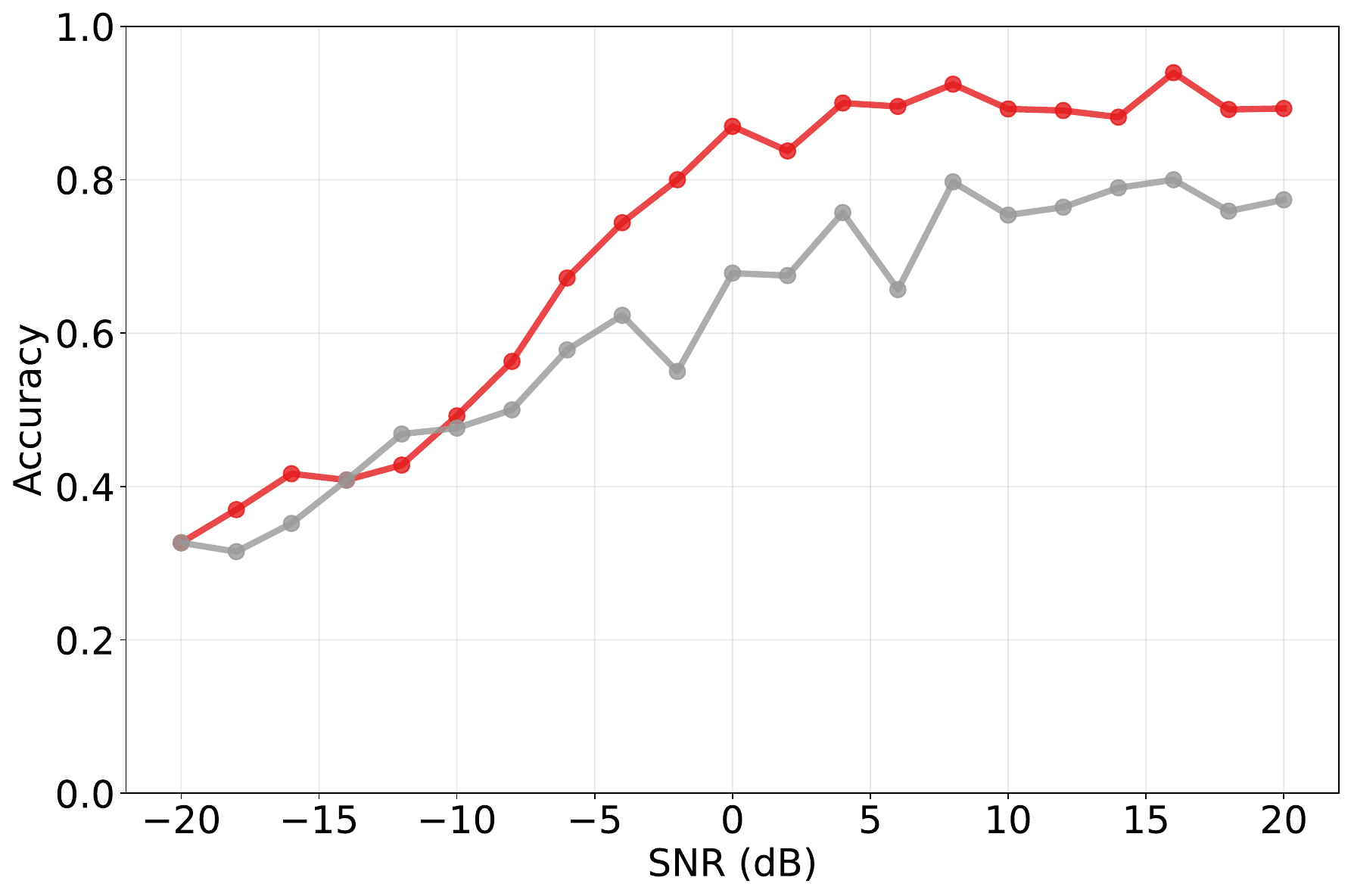}
        \caption{Pulse Width}
    \end{subfigure}
    \hfill
    \begin{subfigure}{0.19\textwidth}
        \includegraphics[width=\linewidth]{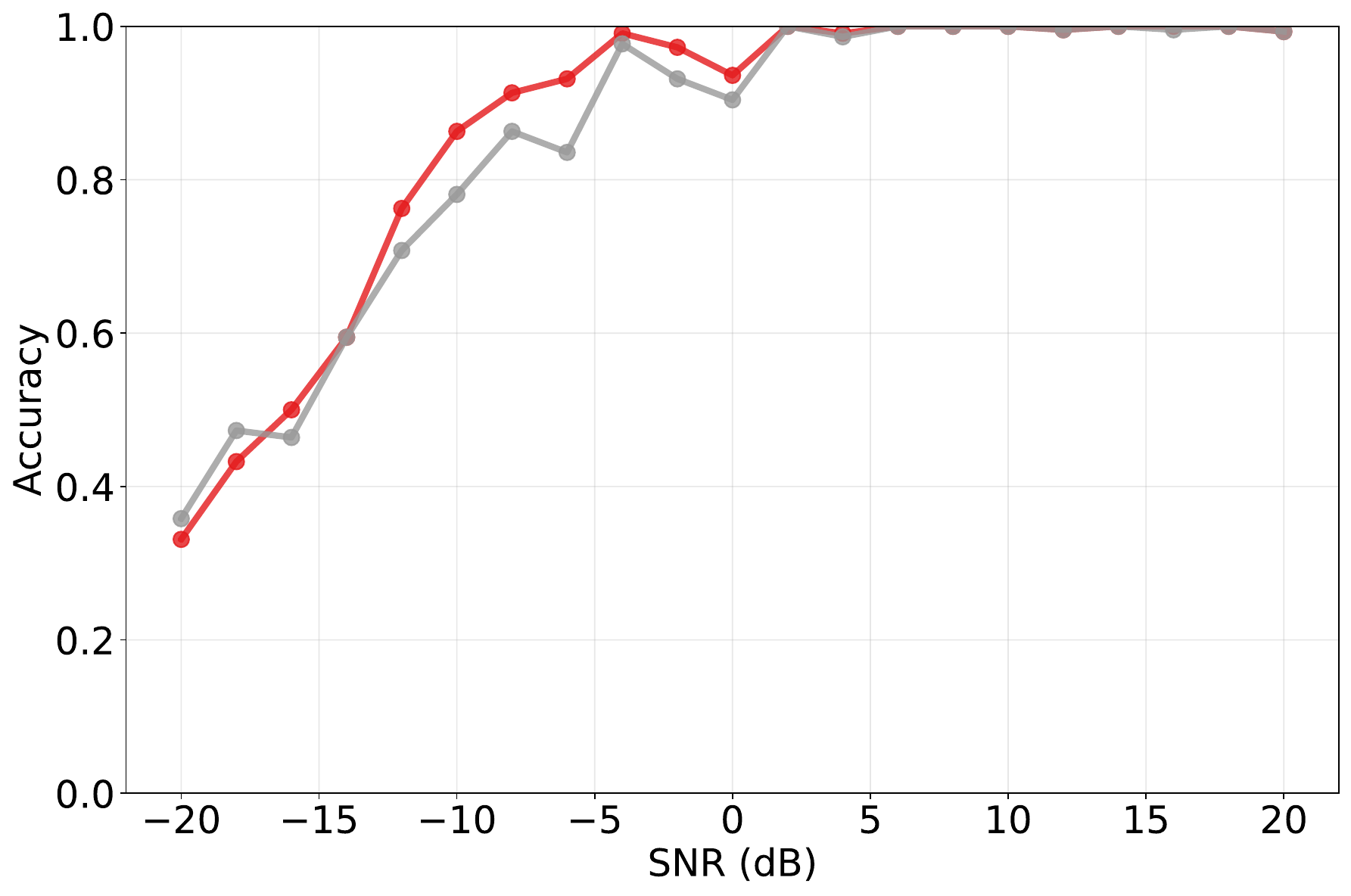}
        \caption{Protocol ID}
    \end{subfigure}
    \hfill
    \begin{subfigure}{0.19\textwidth}
        \includegraphics[width=\linewidth]{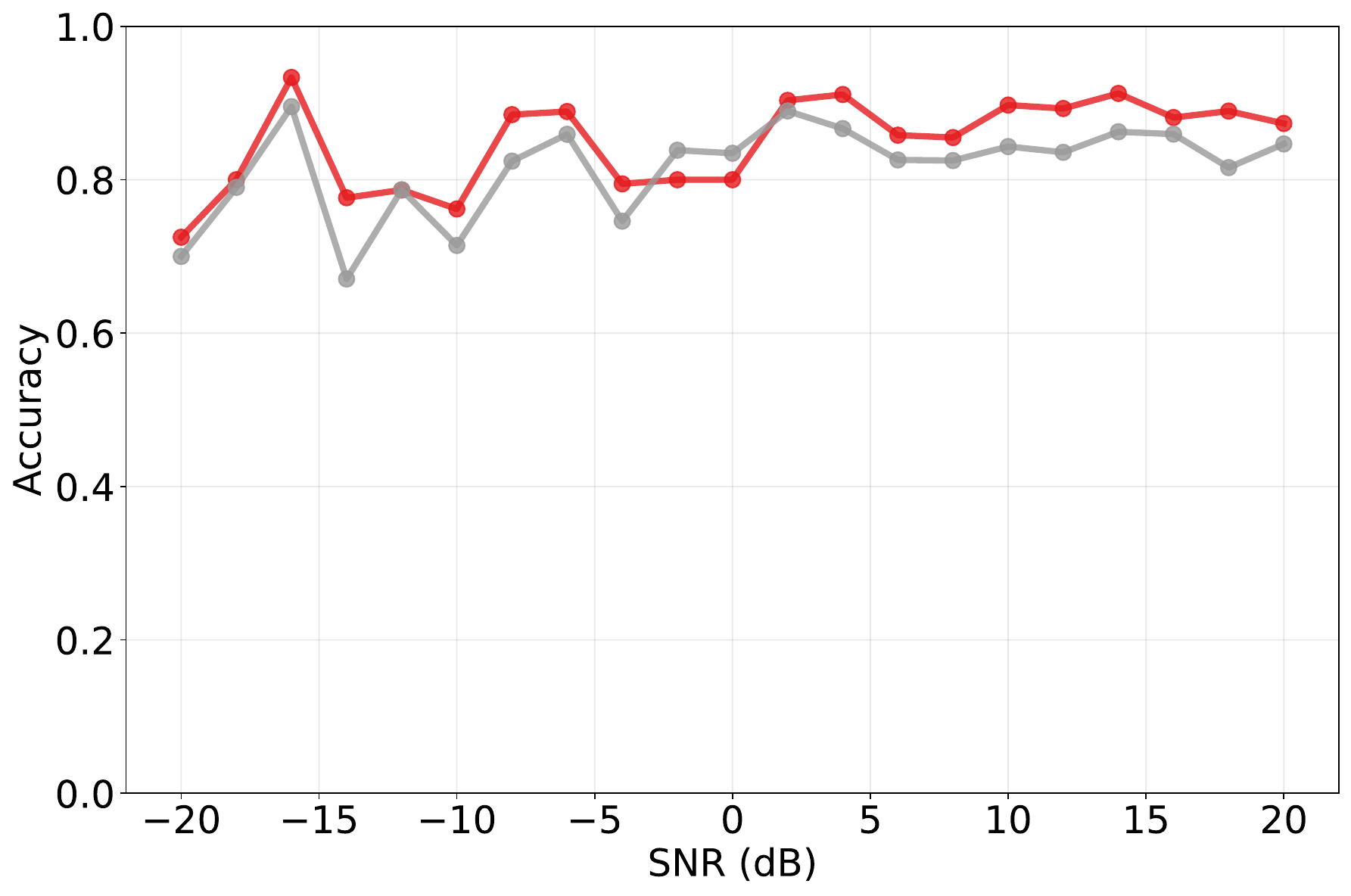}
        \caption{Comm. Jamming}
    \end{subfigure}
    \hfill
    \begin{subfigure}{0.19\textwidth}
        \includegraphics[width=\linewidth]{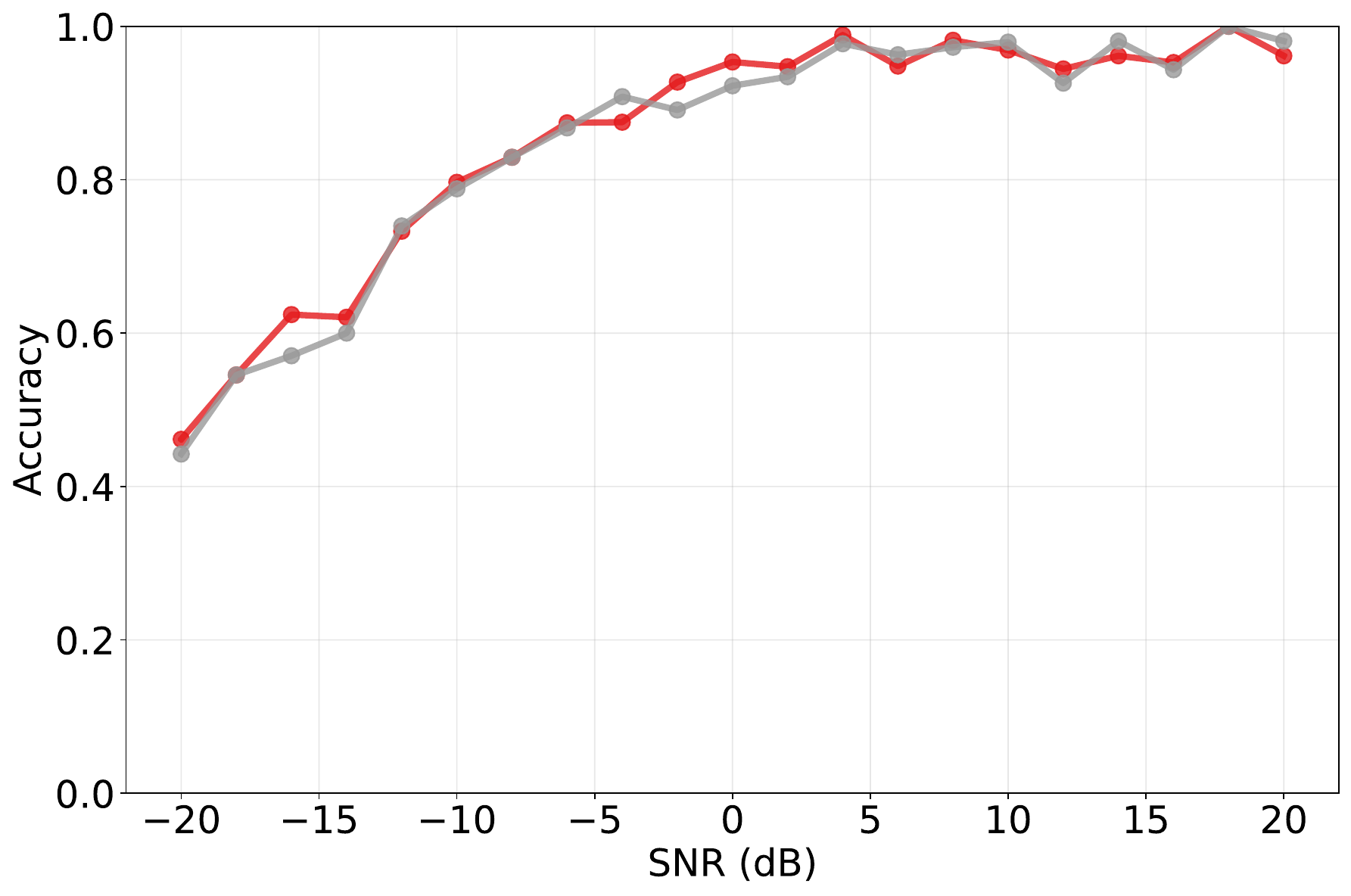}
        \caption{Radar Jamming}
    \end{subfigure}
    \hfill
    \begin{subfigure}{0.19\textwidth}
        \includegraphics[width=\linewidth]{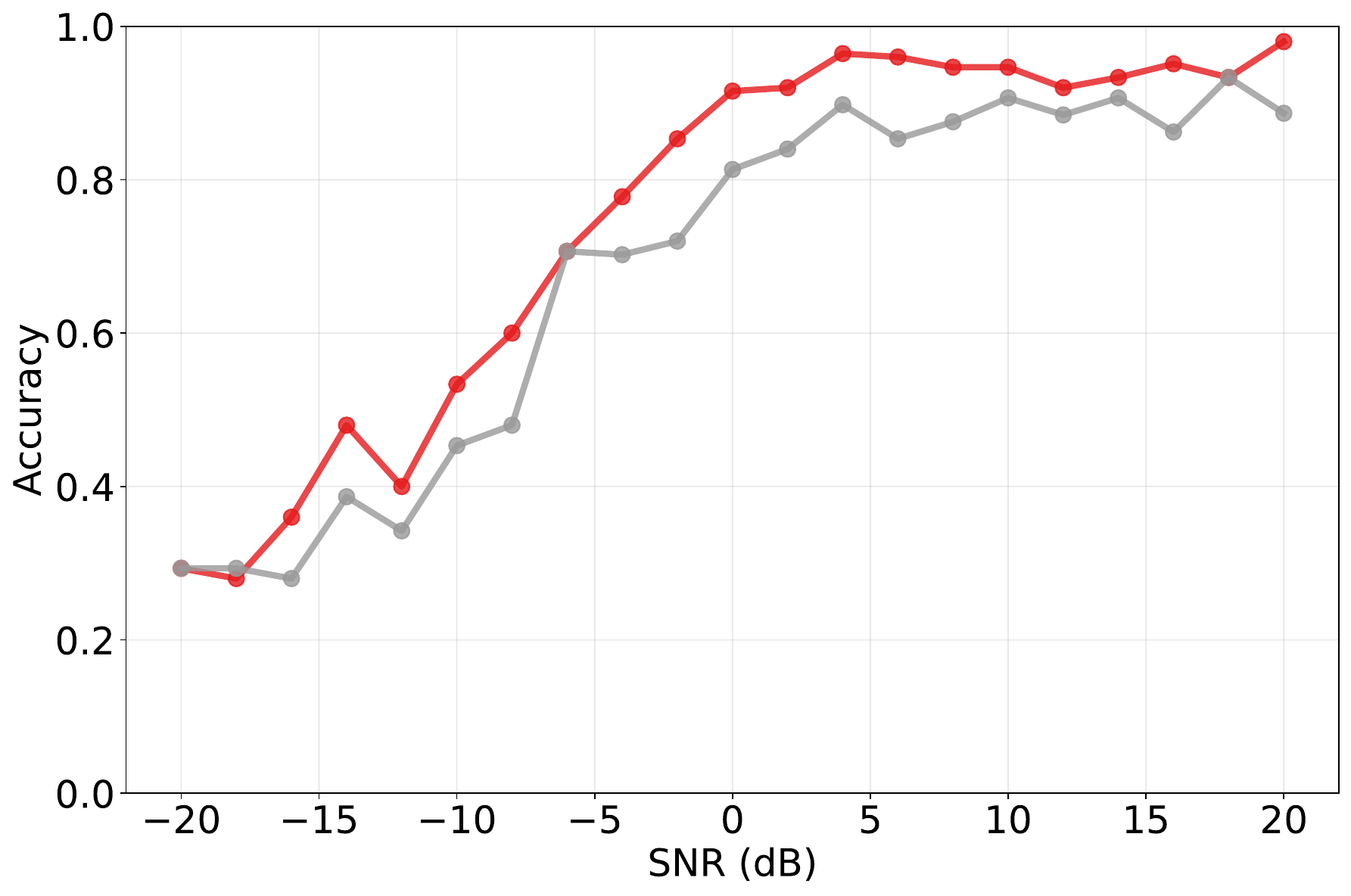}
        \caption{Segment Det.}
    \end{subfigure}
    
    \caption{\textbf{Detailed performance comparison across 10 sub-tasks.} The red line denotes MERLIN and the gray line denotes the Stage-1 baseline. MERLIN demonstrates significant improvements in complex parameter estimation tasks (e.g., Duty Cycle, Pulse Width) across all SNR levels, while maintaining robust performance in classification tasks.}
    \label{fig:snr_breakdown}
\end{figure*}

Figure~\ref{fig:snr_breakdown} provides a comprehensive breakdown of model performance across 10 sub-tasks under varying SNR conditions. The comparison between MERLIN and the Stage-1 baseline reveals distinct behavioral patterns:

\begin{itemize}
    \item \textbf{Superiority in Fine-grained Perception:} In complex parameter estimation tasks such as \textit{Duty Cycle}, \textit{Pulse Width}, and \textit{PRF Estimation}, MERLIN exhibits a consistent and significant performance lead (often $>10\%$) across the entire SNR spectrum. This indicates that our feature-level distillation not only improves noise robustness but also enhances the model's fundamental ability to extract subtle signal features that the baseline fails to capture even at high SNRs.
    
    \item \textbf{Robustness in Modulation and Detection:} For \textit{Modulation Recognition (MOD)} and \textit{Signal Detection (SD)}, MERLIN demonstrates stronger resilience. While both models degrade as noise increases, MERLIN maintains a clearer operational margin in challenging low-SNR environments (e.g., $-10$ dB to $0$ dB).
    
    \item \textbf{Performance on Saturated Tasks:} For tasks like \textit{Protocol Identification (PI)} and \textit{Jamming Recognition (RJR)}, both models achieve high accuracy rapidly. However, MERLIN remains competitive and consistently matches or slightly exceeds the baseline, ensuring no degradation in simpler tasks while significantly boosting performance in complex ones.
\end{itemize}

\clearpage
\subsection{Visualizations of samples and challenging cases}
\vspace{1cm} 
\begin{center}
    \hspace*{0.2cm} 
    \includegraphics[width=0.85\textwidth]{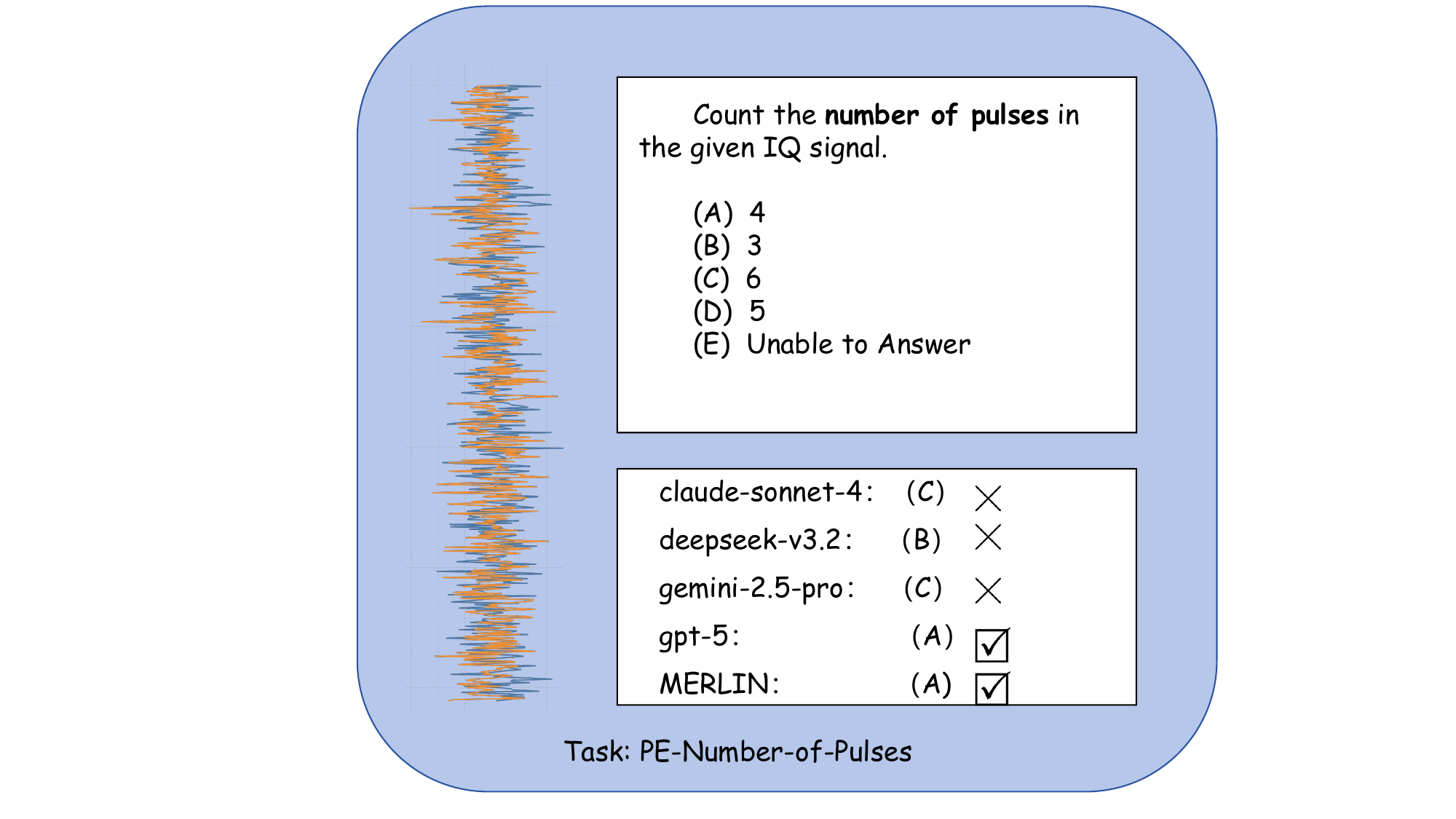}
    \captionof{figure}{Number-of-Pulses estimation details}
    \label{fig:sample1}
    
    \vspace{3cm} 

    \hspace*{-1.2cm} 
    \includegraphics[width=0.85\textwidth]{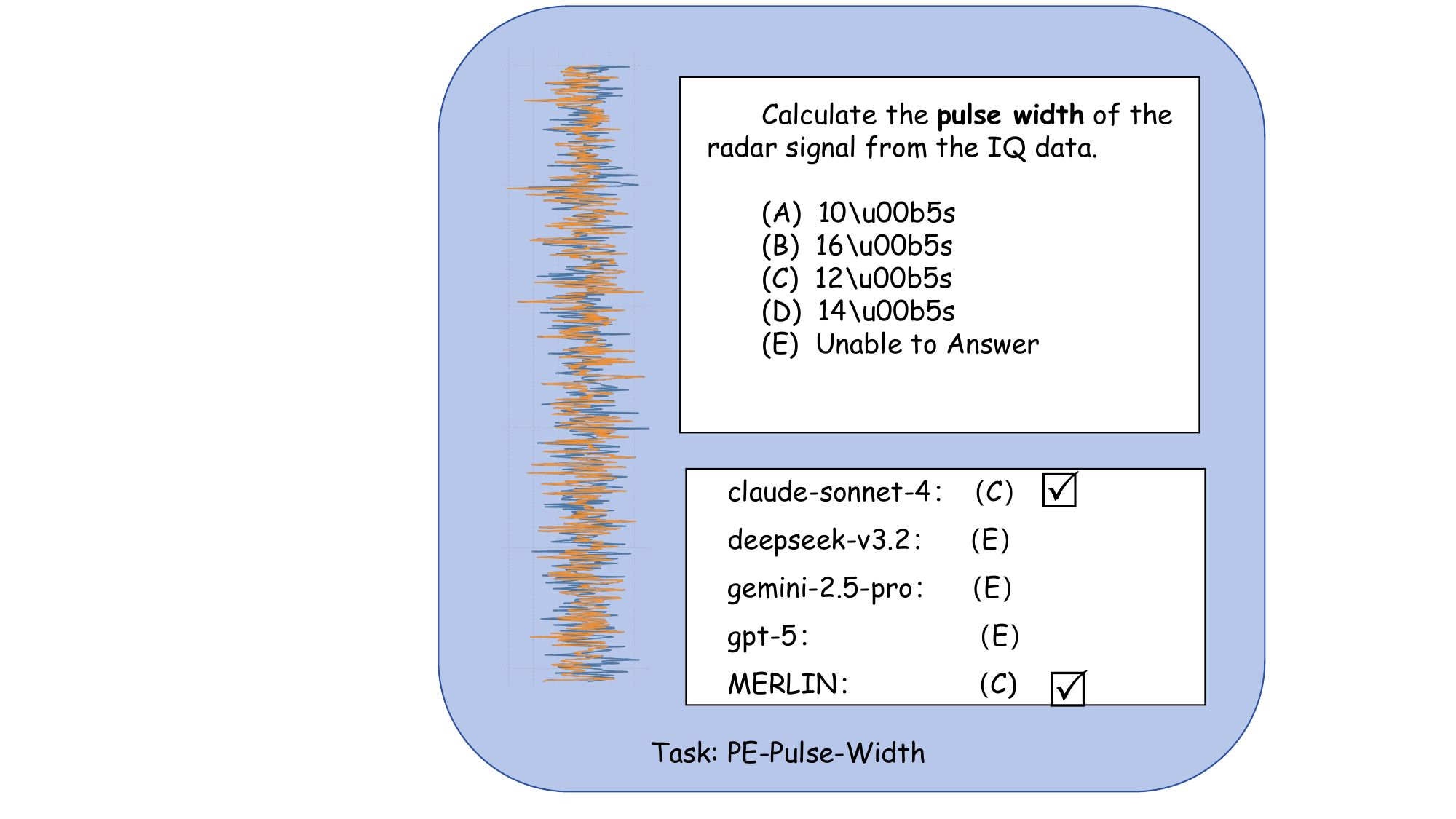}
    \captionof{figure}{ Pulse-Width estimation classification}
    \label{fig:sample2}

    \vspace{1cm} 
    \centering
    \includegraphics[width=0.85\textwidth]{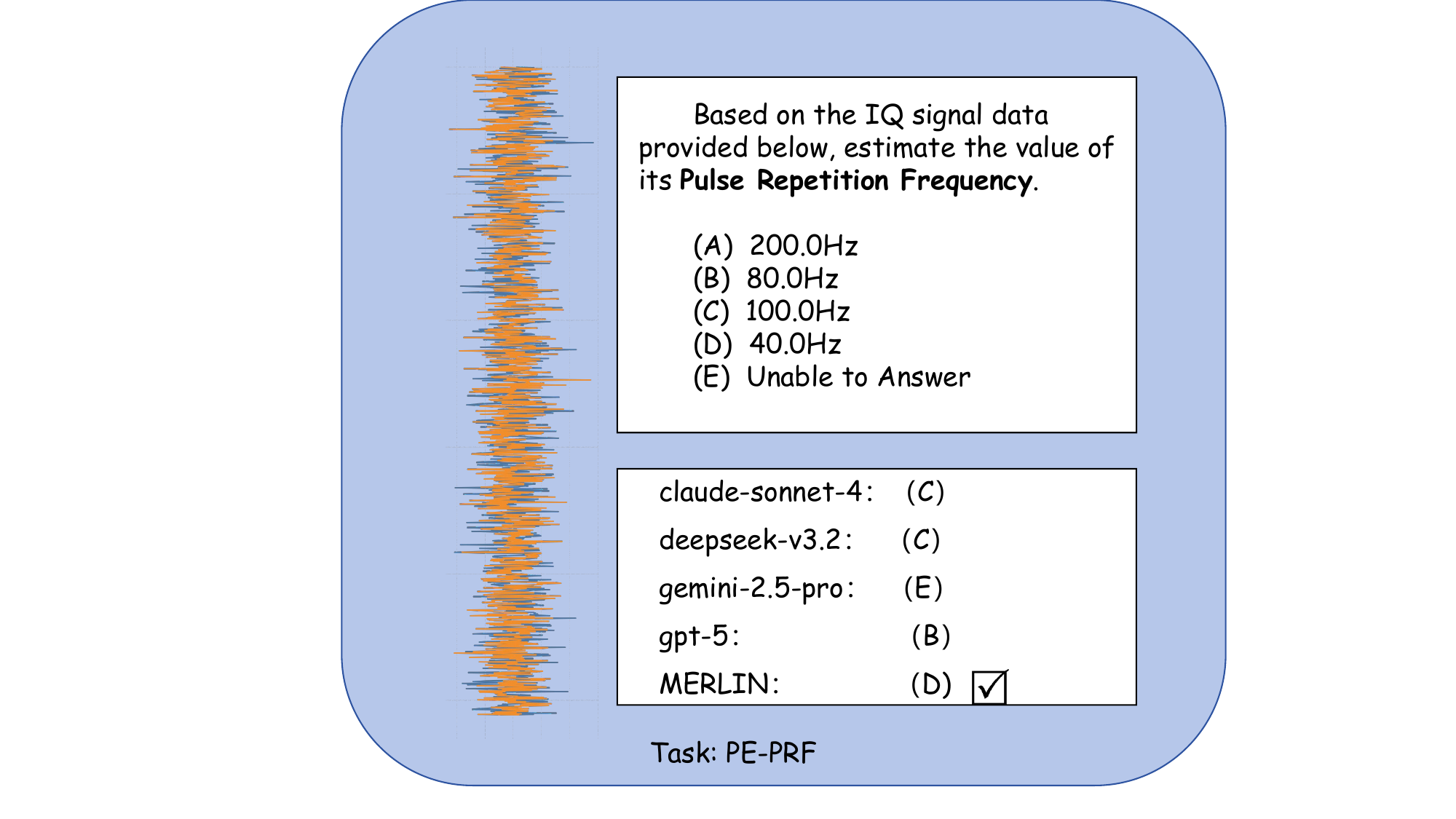}
    \caption{PRF estimation details}
    \label{fig:sample3}
    
    \vspace{3cm} 

    \hspace*{-1.2cm}
    \includegraphics[width=0.85\textwidth]{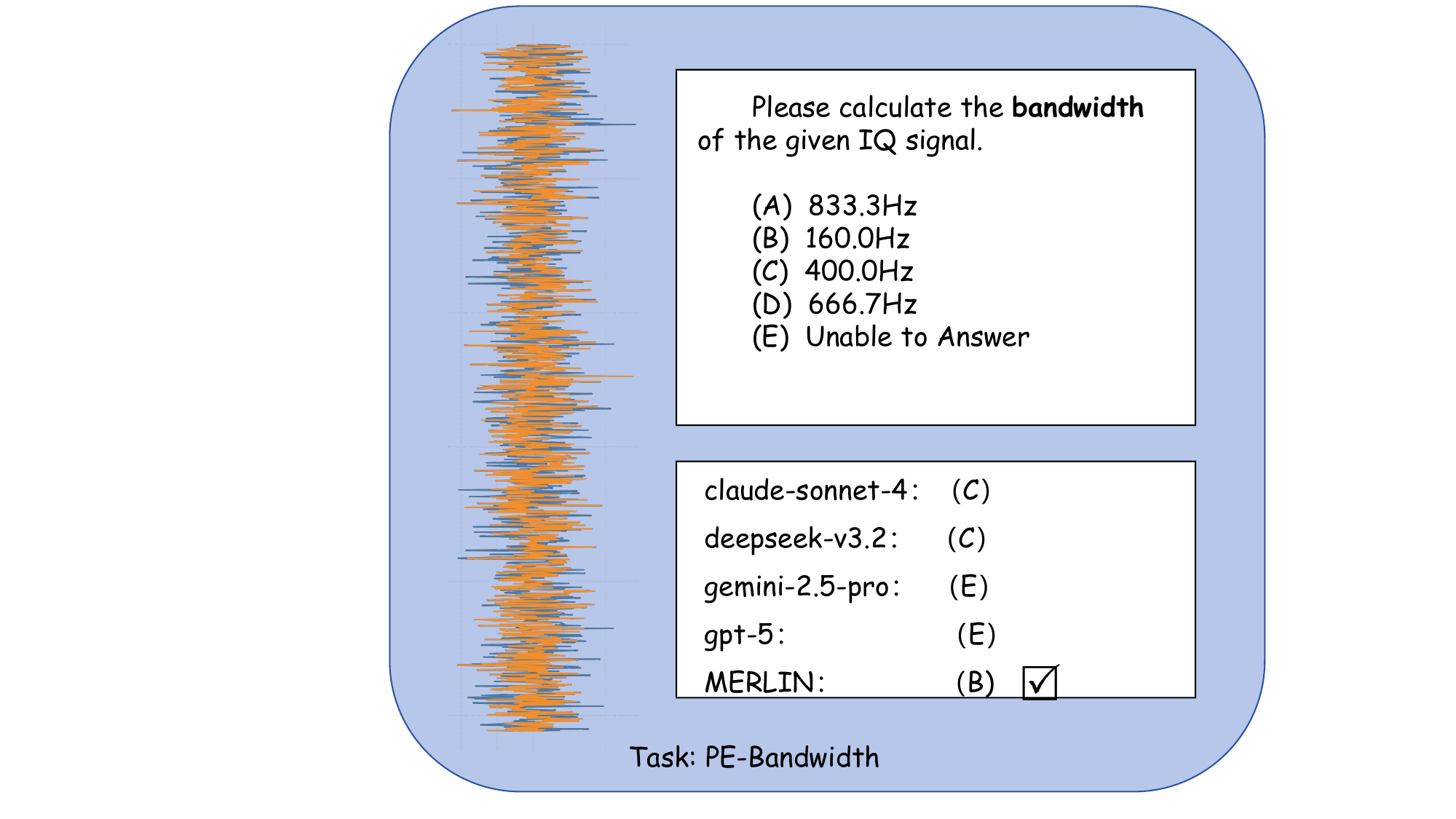}
    \caption{Bandwidth estimation details}
    \label{fig:sample4}

    \vspace{1cm} 
    \hspace*{-0.5cm}
    \includegraphics[width=0.85\textwidth]{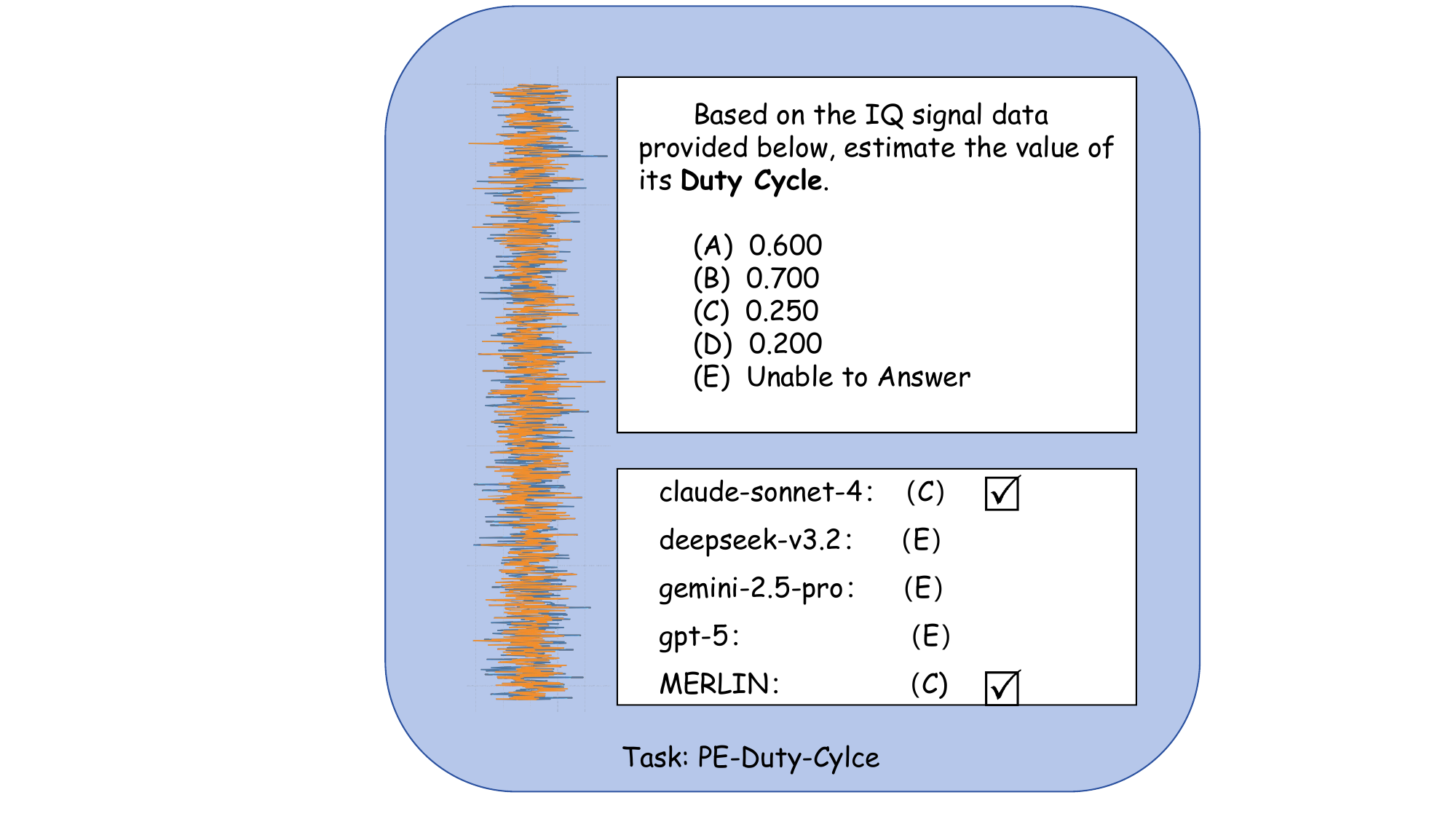}
    \caption{Duty-Cylce estimation details}
    \label{fig:sample5}

    \vspace{3cm} 
    \hspace*{-0.5cm}
    \includegraphics[width=0.85\textwidth]{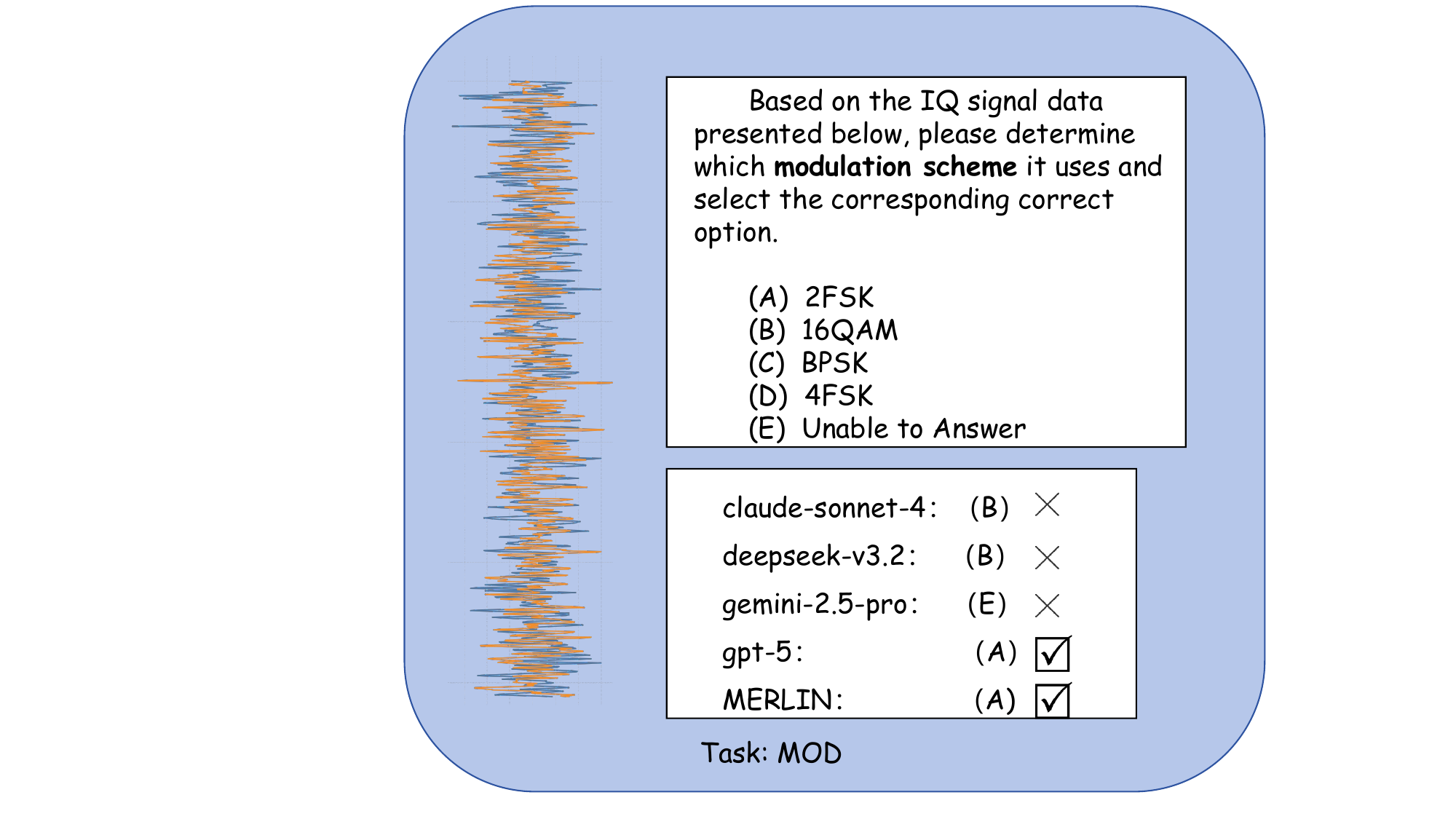}
    \caption{Modulation details}
    \label{fig:sample6}

    \vspace{1cm} 
    \centering
    \includegraphics[width=0.85\textwidth]{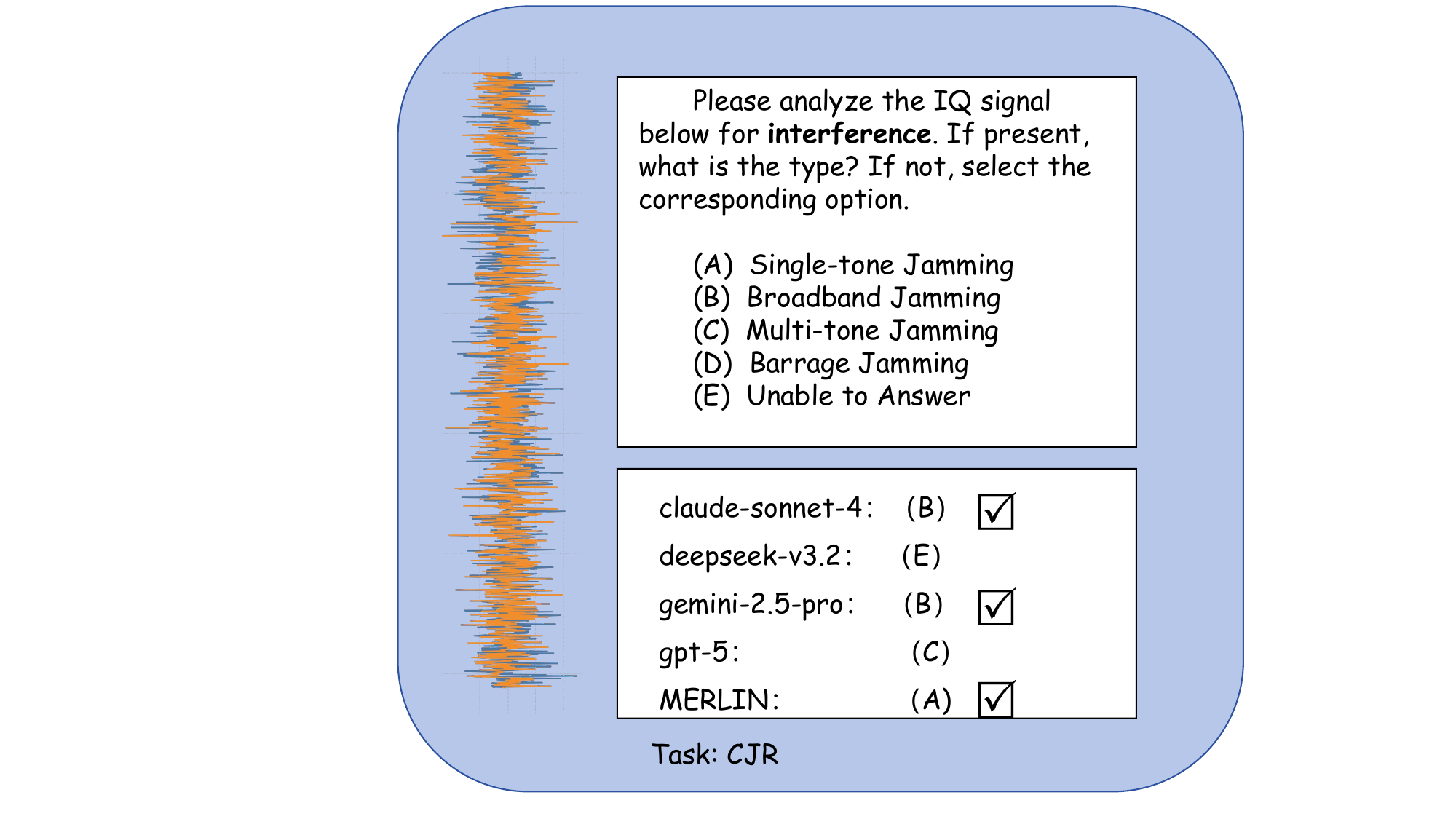}
    \caption{CJR details}
    \label{fig:sample7}

    \vspace{3cm} 
    \hspace*{0.5cm}
    \includegraphics[width=0.85\textwidth]{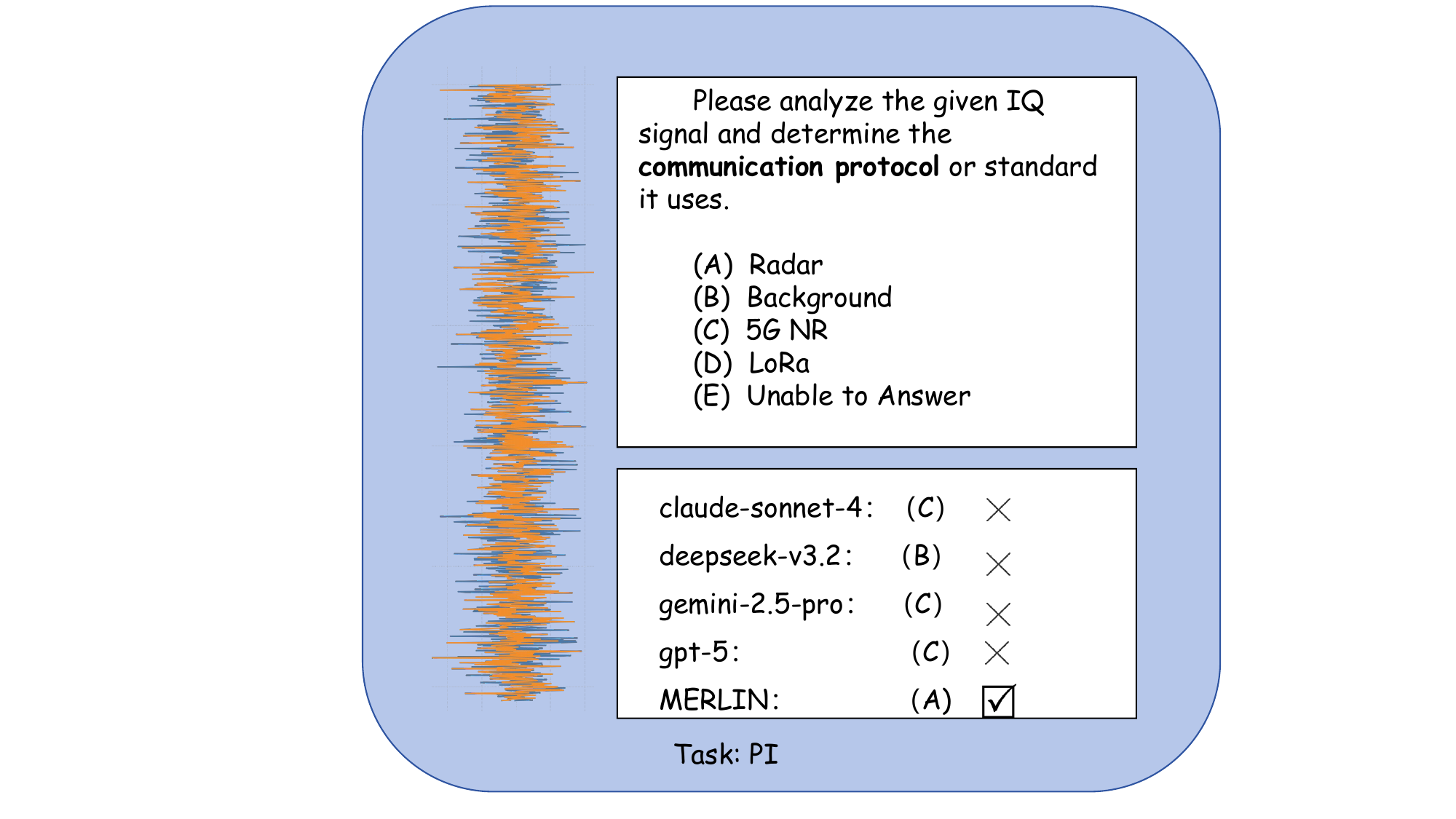}
    \caption{PI details}
    \label{fig:sample8}

    \vspace{1cm} 
    \centering
    \includegraphics[width=0.85\textwidth]{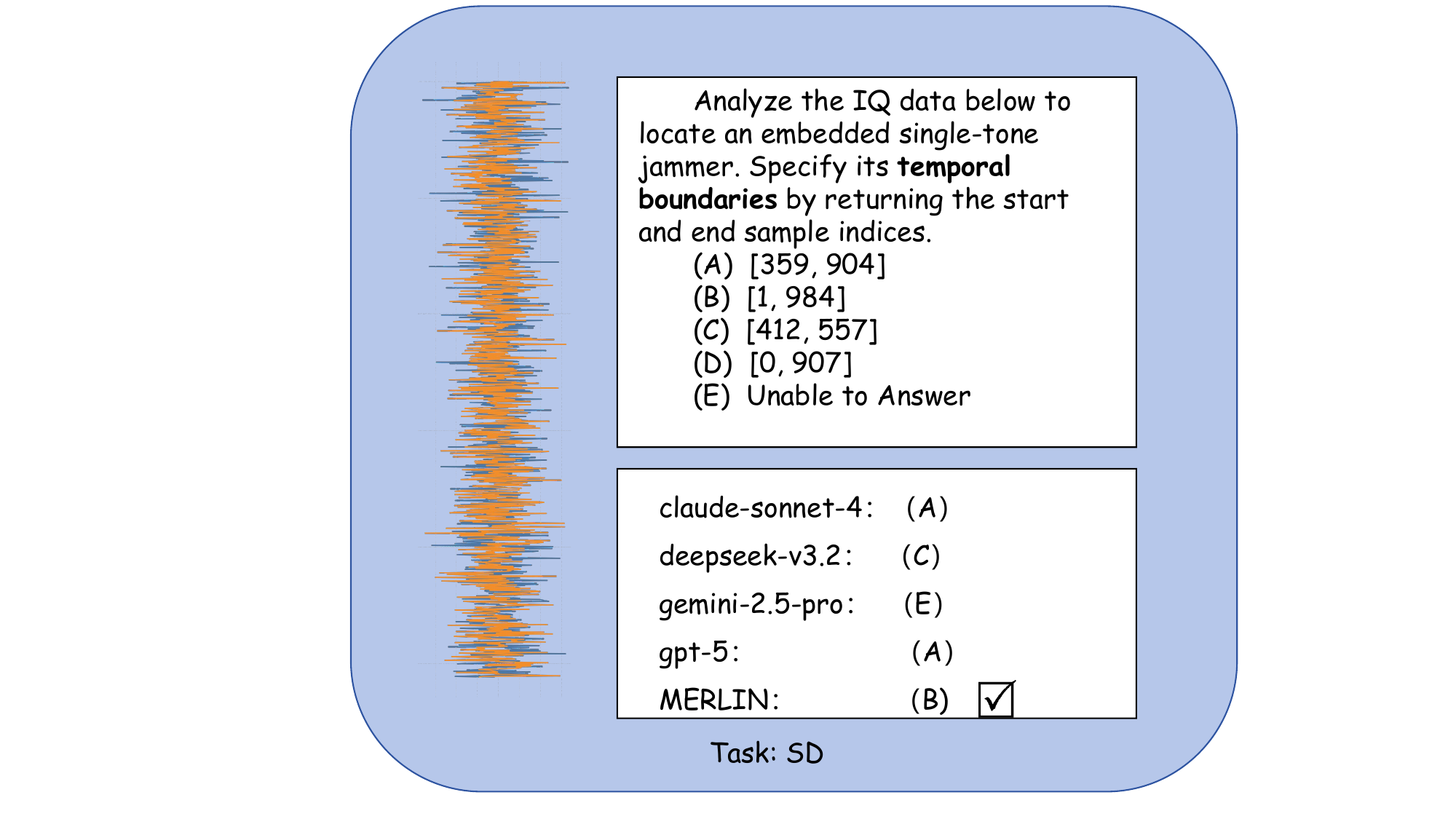}
    \caption{SD details}
    \label{fig:sample9}
    
    \vspace{3cm} 
    \hspace*{-0.3cm}
    \centering
    \includegraphics[width=0.85\textwidth]{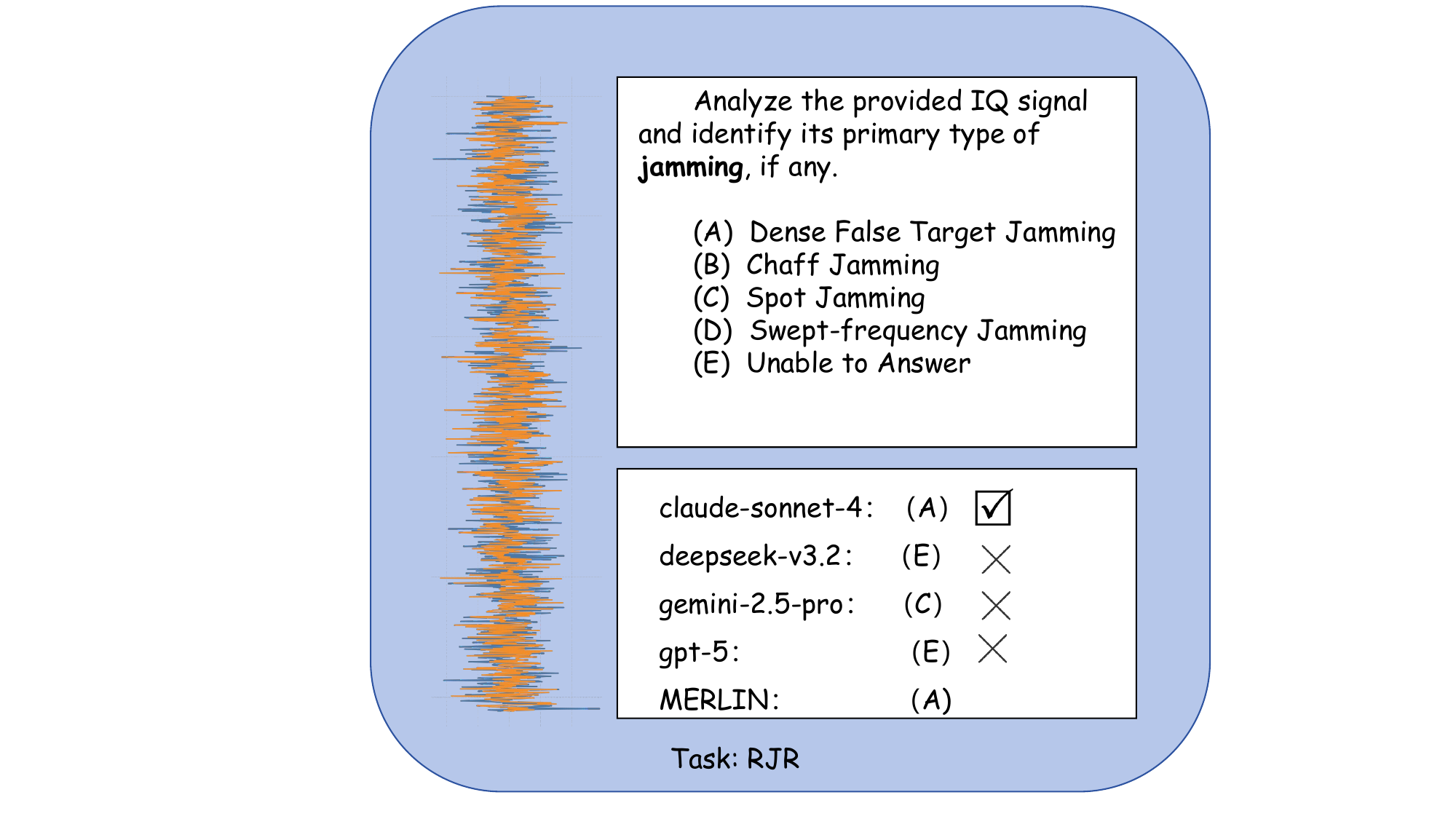}
    \caption{RJR details}
    \label{fig:sample10}
\end{center}


\end{document}